\documentclass[12pt,twoside]{article}
\usepackage{latexsym,amsmath,amssymb,graphicx}
\def\,{\mskip 3mu} \def\>{\mskip 4mu plus 2mu minus 4mu} \def\;{\mskip 5mu plus 5mu} \def\!{\mskip-3mu}
\def\dispmuskip{\thinmuskip= 3mu plus 0mu minus 2mu \medmuskip=  4mu plus 2mu minus 2mu \thickmuskip=5mu plus 5mu minus 2mu}
\def\textmuskip{\thinmuskip= 0mu                    \medmuskip=  1mu plus 1mu minus 1mu \thickmuskip=2mu plus 3mu minus 1mu}
\textmuskip
\def\beq{\dispmuskip\begin{equation}}    \def\eeq{\end{equation}\textmuskip}
\def\beqn{\dispmuskip\begin{displaymath}}\def\eeqn{\end{displaymath}\textmuskip}
\def\bqa{\dispmuskip\begin{eqnarray}}    \def\eqa{\end{eqnarray}\textmuskip}
\def\bqan{\dispmuskip\begin{eqnarray*}}  \def\eqan{\end{eqnarray*}\textmuskip}

\newtheorem{theorem}{Theorem}

\newtheorem{lemma}[theorem]{Lemma}
\newtheorem{definition}[theorem]{Definition}

\newtheorem{proposition}[theorem]{Proposition}
\newtheorem{principle}[theorem]{Principle}
\newenvironment{keywords}{\centerline{\bf\small
Keywords}\begin{quote}\small}{\par\end{quote}\vskip 1ex}

\def\citet{\cite}
\def\citep{\cite}
\def\citeauthor{\cite}
\newtheorem{myexample}[theorem]{Example}
\def\fexample#1#2#3{\vspace{-1ex}\begin{myexample}[#2]\label{#1}\rm #3
\hspace*{\fill} $\diamondsuit\quad$ \end{myexample}\vspace{-2ex} }
\def\paradot#1{\vspace{1.3ex plus 0.7ex minus 0.5ex}\noindent{\bf\boldmath{#1.}}}

\def\eps{\varepsilon}
\def\nq{\hspace{-1em}}
\def\qed{\hspace*{\fill}\rule{1.4ex}{1.4ex}$\quad$\\}
\def\fr#1#2{{\textstyle{#1\over#2}}}
\def\frs#1#2{{^{#1}\!/\!_{#2}}}
\def\SetR{I\!\!R}
\def\SetN{I\!\!N}

\def\SetZ{Z\!\!\!Z}
\def\qmbox#1{{\quad\mbox{#1}\quad}}
\def\e{{\rm e}}                        % natural e

\def\E{{\bf E}}                         % Expectation
\def\P{{\rm P}}                         % Probability
\def\I{1\!\!1}        % identity matrix
\def\lb{{\log_2}}
\def\v{\boldsymbol}
\def\trp{{\!\top\!}}

\def\G{\Gamma}
\def\a{\alpha}

\def\g{\gamma}
\def\s{\sigma}
\def\t{\theta}
\def\b{\beta}
\def\l{\lambda}
\def\r{\rho}
\def\D{{\cal D}}
\def\F{{\cal F}}
\def\M{{\cal M}}
\def\S{{\cal S}}
\def\N{{\cal N}}
\def\R{{\cal R}}
\def\X{{\cal X}}
\def\Y{{\cal Y}}
\def\L{\text{\rm Loss}}
\def\h{\hat}
\def\Rank{\text{\rm Rank}}
\def\LR{\text{\rm LR}}
\def\tr{\text{\rm tr}}
\def\Gauss{\text{Gauss}}
\def\KL{\text{\rm KL}}

\def\MDL{\text{\rm MDL}}

\def\GCV{\text{\rm GCV}}

\def\arg{\text{\rm arg}}
\def\SNR{\text{\rm SNR}}

\begin{document}
%%%%%%%%%%%%%%%%%%%%%%%%%%%%%%%%%%%%%%%%%%%%%%%%%%%%%%%%%%%%%%%
%%                    T i t l e - P a g e                    %%
%%%%%%%%%%%%%%%%%%%%%%%%%%%%%%%%%%%%%%%%%%%%%%%%%%%%%%%%%%%%%%%

\title{\vspace{-2ex}
\bf\Large\hrule height5pt \vskip 4mm
Model Selection with the Loss Rank Principle
\vskip 2mm \hrule height2pt}

\author{
{\bf Marcus Hutter}\\[1mm]
\normalsize RSISE$\,$@$\,$ANU and SML$\,$@$\,$NICTA,
\normalsize Canberra, ACT, 0200, Australia \\[-1mm]
\normalsize \texttt{marcus@hutter1.net \ \ \ \ www.hutter1.net}
\and
{\bf Minh-Ngoc Tran}\\[1mm]
\normalsize Department of Statistics and Applied Probability,  \\[-1mm]
\normalsize National University of Singapore, \ \ \ \texttt{ngoctm@nus.edu.sg} }

\date{2 March 2010}

\maketitle

\vspace*{-5ex}\begin{abstract}\noindent
A key issue in statistics and machine learning is to automatically
select the ``right'' model complexity, e.g., the number of neighbors
to be averaged over in k nearest neighbor (kNN) regression or the
polynomial degree in regression with polynomials. We suggest a novel
principle - the Loss Rank Principle (LoRP) - for model selection in regression and
classification. It is based on the loss rank, which counts how many
other (fictitious) data would be fitted better. LoRP selects the
model that has minimal loss rank.
Unlike most penalized maximum likelihood variants (AIC, BIC, MDL),
LoRP depends only on the regression functions
and the loss function. It works without a stochastic noise model,
and is directly applicable to any non-parametric regressor, like
kNN.
\def\contentsname{\centering\normalsize Contents}
{\parskip=-2.7ex\tableofcontents}
\end{abstract}

\begin{keywords}\vspace{-1ex}
Model selection,
loss rank principle,
non-parametric regression,
classification,
general loss function,
k nearest neighbors.
\end{keywords}

\newpage
%%%%%%%%%%%%%%%%%%%%%%%%%%%%%%%%%%%%%%%%%%%%%%%%%%%%%%%%%%%%%%%
\section{Introduction}\label{secIntro}
%%%%%%%%%%%%%%%%%%%%%%%%%%%%%%%%%%%%%%%%%%%%%%%%%%%%%%%%%%%%%%%

%-------------------------------%
\paradot{Regression}
%-------------------------------%
Consider a regression or classification problem in which we want to
determine the functional relationship $y_i\approx f_{\text{true}}(x_i)$
from data $D = \{(x_1,y_1),...,(x_n,y_n)\}$, i.e., we seek a
function $r(.|D)\equiv r(D)(.)$ such that $r(x|D)\equiv r(D)(x)$ is close to the unknown
$f_{\text{true}}(x)$ for all $x$. One may define $r(.|D)$ directly, e.g.,\
``average the $y$ values of the $k$ nearest neighbors (kNN) of $x$
in $D$", or select $r(.|D)$ from a class of functions $\F$ that has
smallest (training) error on $D$. If the class $\F$ is not too
large, e.g., the polynomials of fixed reasonable degree $d$, this
often works well.

%-------------------------------%
\paradot{Model selection}
%-------------------------------%
What remains is to select the right model complexity $c$, like $k$
or $d$. This selection cannot be based on the training error, since
the more complex the model (large $d$, small $k$) the better the fit
on $D$ (perfect for $d=n$ and $k=1$). This problem is called
overfitting, for which various remedies have been suggested.

The most popular ones in practice are based on a test set $D'$ used
for selecting the $c$ for which the function $r_c(.|D)$ has smallest
(test) error on $D'$, or improved versions like cross-validation
\citep{All:74}. Typically $D'$ is cut from $D$, thus
reducing the sample size available for regression. Test set methods
often work well in practice, but the reduced sample decreases
accuracy, which can be a serious problem if $n$ is small. We will
not discuss empirical test set methods any further. See
\cite{MacKay:92} for a comparison of cross-validation with Bayesian
model selection.

There are also various model selection methods that allow to use all
data $D$ for regression. The most popular ones can be regarded as
penalized versions of Maximum Likelihood (ML). In addition to the
function class $\F_c$ (subscript $c$ belonging to some set indexing
the complexity), one has to specify a sampling model $\P(D|f)$,
e.g.,\ that the $y_i$ have independent Gaussian distribution with
mean $f(x_i)$. ML chooses $r_c(D)=\arg\max_{f\in\F_c}\P(D|f)$,
Penalized ML (PML) then chooses $\h
c=\arg\min_c\{-\log\P(D|r_c(D))+$Penalty$(c)\}$, where the penalty
depends on the used approach (MDL \citep{Rissanen:78}, BIC
\citep{Schwarz:78}, AIC \citep{Akaike:73}). All PML variants rely on
a proper sampling model (which may be difficult to establish),
ignore (or at least do not tell how to incorporate) a potentially
given loss function (see \cite{Yamanishi:99,Gruenwald:04} for
exceptions), are based on distribution-independent penalties (which
may result in bad performance for specific distributions), and
are typically limited to (semi)parametric models.

%-------------------------------%
\paradot{Main idea}
%-------------------------------%
The main goal of the paper is to establish a criterion for selecting
the ``best'' model complexity $c$ %
based on regressors $r_c$ given as a black box without insight into the
origin or inner structure of $r_c$, %
that does not depend on things often not given (like a stochastic noise model), %
and that exploits what is/should be given (like the loss function,
note that the criterion can also be used for loss-function selection, see Section \ref{secLFS}).
The key observation we exploit is that large classes $\F_c$ or more
flexible regressors $r_c$ can fit more data well than
more rigid ones. We define the {\em loss rank} of $r_c$ as the number of
other (fictitious) data $D'$ that are fitted better by
$r_c(D')$ than $D$ is fitted by $r_c(D)$, as measured by some loss
function.
The loss rank is large for regressors fitting $D$ not well {\em and} for
too flexible regressors (in both cases the regressor fits many other
$D'$ better). The loss rank has a minimum for not too flexible
regressors which fit $D$ not too bad. We claim that minimizing
the loss rank is a suitable model selection criterion, since it
trades off the quality of fit with the flexibility of the model.
Unlike PML, our Loss Rank Principle (LoRP) works without a noise
(stochastic sampling) model, and is directly applicable to any
non-parametric regressor, like kNN.

%-------------------------------%
\paradot{Related ideas}
%-------------------------------%
There are various other ideas that somehow count
fictitious data.
In normalized ML \citep{Gruenwald:04}, the
complexity of a stochastic model class is defined as the log sum
over all $D'$ of maximum likelihood probabilities.
In the luckiness framework for classification \cite[Chp.4]{Herbrich:02}, the
loss rank is related to the level of a hypothesis, if the
empirical loss is used as an unluckiness function.
The empirical Rademacher complexity \citep{Kol:01,BBL:02} averages over all possible relabeled instances.
Finally, instead of considering all $D'$ one could
consider only the set of all permutations of $\{y_1,...,y_n\}$, like
in permutation tests \citep{Efron:93}. The test statistic would here be the empirical
loss.

%-------------------------------%
\paradot{Contents}
%-------------------------------%
In Section \ref{secLoRP}, after giving a brief introduction to
regression, we formally state LoRP for model selection.
Explicit expressions for the loss rank for the important class of
linear regressors are derived in Section \ref{secLM}; this class
includes kNN, polynomial, linear basis function (LBFR), kernel,
projective regression, and some others.
In Section \ref{secOpt}, we establish optimality properties of LoRP
for linear regression, namely model consistency and asymptotic mean
efficiency.
Experiments are presented in Section \ref{secExp}: We compare
LoRP to other selection methods and demonstrate the use of LoRP for some specific problems
like choosing tuning parameters in kNN and spline regression.
In Section \ref{secBayes} we compare linear LoRP to Bayesian model
selection for linear regression with Gaussian noise and prior,
and in Section \ref{secOMS} to PML, in particular MDL, BIC, and AIC,
and then discuss two trace formulas for the effective dimension.
Sections \ref{secLFS}-\ref{secKNN} can be considered as extension sections.
In Section \ref{secLFS} we show how to generalize linear LoRP to
non-quadratic loss, in particular to other norms. We also discuss
how LoRP can be used to select the loss function itself, in case it
is not part of the problem specification.
In Section \ref{secSCR} we briefly discuss interpolation. LoRP only
depends on the regressor on data $D$ and not on
$x\not\in\{x_1,...,x_n\}$. We construct canonical regressors for
off-data interpolation from regressors given only on-data, in
particular for kNN, Kernel, and LBFR, and show that they are
canonical.
In Section \ref{secKNN} we derive exact expressions for kNN when
$\{x_1,...,x_n\}$ forms a discrete $d$-dimensional hypercube, and
discuss the limits $n\to\infty$, $k\to\infty$, and $d\to\infty$.
Section \ref{secMisc} contains the conclusions of our work and
further considerations that could be elaborated on in the future.

The main idea of LoRP has already been presented at the COLT 2007 conference \citep{Hutter:07lorp}.
In this paper we present LoRP more thoroughly, discover its theoretical properties
and evaluate the method through some experiments.

%%%%%%%%%%%%%%%%%%%%%%%%%%%%%%%%%%%%%%%%%%%%%%%%%%%%%%%%%%%%%%%
\section{The Loss Rank Principle}\label{secReg}\label{secLoRP}
%%%%%%%%%%%%%%%%%%%%%%%%%%%%%%%%%%%%%%%%%%%%%%%%%%%%%%%%%%%%%%%

After giving a brief introduction to regression, classification,
model selection, overfitting, and some reoccurring examples,
we state our novel Loss Rank Principle for model
selection. We first state it for classification (Principle
\ref{prLRD} for discrete values), and then generalize it for
regression (Principle \ref{prLRC} for continuous values), and exemplify
it on two (over-simplistic) artificial Examples \ref{exSD} and
\ref{exSC}. Thereafter we show how to regularize LoRP for realistic
regression problems.

%-------------------------------%
\paradot{Setup and notation}
%-------------------------------%
We assume data $D = (\v x,\v y) := \{(x_1,y_1),...,(x_n,y_n)\} \in
(\X\times\Y)^n=:\D$ has been observed. We think of the $y$ as having an
approximate functional dependence on $x$, i.e., $y_i\approx
f_{\text{true}}(x_i)$, where $\approx$ means that the $y_i$ are distorted
by noise from the unknown ``true'' values
$f_{\text{true}}(x_i)$.
We will write $(x,y)$ for generic data points, use vector notation
$\v x=(x_1,...,x_n)^\trp$ and $\v y=(y_1,...,y_n)^\trp$, and $D'=(\v
x',\v y')$ for generic (fictitious) data of size $n$. A full list of
abbreviations and notations used throughout the paper is placed in
the appendix.

%-------------------------------%
\paradot{Regression and classification}
%-------------------------------%
In regression problems $\Y$ is typically (a subset of) the real
set $\SetR$ or some more general measurable space like
$\SetR^m$. In classification, $\Y$ is a finite set or at least
discrete. We impose no restrictions on $\X$. Indeed, $\v x$ will
essentially be fixed and plays only a spectator role, so we will
often notationally suppress dependencies on $\v x$.
The goal of regression/classification is to find a function
$f_D\in\F\subset\X\to\Y$ ``close'' to $f_{\text{true}}$ based on the past
observations $D$. Or phrased in another way: we are interested in a
mapping $r:\D\to\F$ such that $\h y:=r(x|D)\equiv
r(D)(x)\equiv f_D(x)\approx f_{\text{true}}(x)$ for all $x\in\X$.

%-------------------------------%
\fexample{exPR}{polynomial regression}{
%-------------------------------%
For $\X=\Y=\SetR$, consider the set $\F_d:=\{f_{\v w}(x)=w_d
x^{d-1}+...+w_2 x+w_1 : \v w\in\SetR^d \}$ of polynomials of degree
$d-1$. Fitting the polynomial to data $D$, e.g.,\ by least squares
regression, we estimate $\v w$ with $\v{\h w}_D$. The regression
function $\h y=r_d(x|D)=f_{\v{\h w}_D}(x)$ can be written down in
closed form (see Example \ref{exLBFR}).
} % End of Example

%-------------------------------%
\fexample{exKNN}{k nearest neighbors}{
%-------------------------------%
Let $\Y$ be some vector space like $\SetR$ and $\X$ be a metric
space like $\SetR^m$ with some (e.g.,\ Euclidean) metric
$d(\cdot,\cdot)$. kNN estimates $f_{\text{true}}(x)$ by averaging the
$y$ values of the $k$ nearest neighbors ${\N}_k(x)$ of $x$ in $D$, i.e., $r_k(x|D)={1\over k}\sum_{i\in\N_k(x)} y_i$ %
with $|\N_k(x)|=k$ such that $d(x,x_i)\leq d(x,x_j)$ %
for all $i\in\N_k(x)$ and $j\not\in\N_k(x)$.
} % End of Example

%-------------------------------%
\paradot{Parametric versus non-parametric regression}
%-------------------------------%
Polynomial regression is an example of parametric regression in the
sense that $r_d(D)$ is the optimal function from a family of
functions $\F_d$ indexed by $d<\infty$ real parameters ($\v w$). In
contrast, the kNN regressor $r_k$ is directly given and is not based
on a finite-dimensional family of functions. In general, $r$ may be
given either directly or be the result of an optimization process.

%-------------------------------%
\paradot{Loss function}
%-------------------------------%
The quality of fit to the data is usually measured by a loss function
$\L(\v y,\v{\h y})$, where $\h y_i=\h f_D(x_i)$ is an estimate of $y_i$.
Often the loss is additive: $\L(\v y,\v{\h y})=\sum_{i=1}^n\L(y_i,\h
y_i)$. If the class $\F$ is not too large, good regression functions $r(D)$ can
be found by minimizing the loss w.r.t.\ all $f\in\F$. For instance,
$r_d(D)=\arg\min_{f\in\F_d}\sum_{i=1}^n (y_i-f(x_i))^2$ and $\h
y=r_d(x|D)$ in Example
\ref{exPR}.

%-------------------------------%
\paradot{Regression class and loss}
%-------------------------------%
In the following we assume a class of
regressors $\R$ (whatever their origin), e.g.,\ the kNN regressors
$\{r_k:k\in\SetN\}$ or the least squares polynomial
regressors $\{r_d:d\in\SetN_0:=\SetN\cup\{0\}\}$.
Each regressor $r$ can be thought of as a model.
Throughout the paper, we use the terms ``regressor" and ``model" interchangeably.
Note that unlike $f\in\F$,
regressors $r\in\R$ are not functions of $x$ alone but depend
on all observations $D$, in particular on $\v y$.
Like for functions $f$, we can compute the empirical loss of each regressor
$r\in\R$:
\beqn
  \L_r(D) \;\equiv\; \L_r(\v y|\v x) \;:=\; \L(\v y,\v{\h y})
  \;=\; \sum_{i=1}^n \L(y_i,r(x_i|\v x,\v y))
\eeqn
where $\h y_i=r(x_i|D)$ in the third expression, and the last
expression holds in case of additive loss.

%-------------------------------%
\paradot{Overfitting}
%-------------------------------%
Unfortunately, minimizing $\L_r$ w.r.t.\ $r$ will typically {\em
not} select the ``best'' overall regressor. This is the well-known
overfitting problem. In case of polynomials, the classes
$\F_d\subset\F_{d+1}$ are nested, hence $\L_{r_d}$ is monotone
decreasing in $d$ with $\L_{r_n}\equiv 0$ perfectly fitting the
data. In case of kNN, $\L_{r_k}$ is more or less an increasing
function in $k$ with perfect regression on $D$ for $k=1$, since no
averaging takes place.
In general, $\R$ is often indexed by a ``flexibility'' or smoothness
or complexity parameter, which has to be properly determined.
The more flexible $r$ is, the closer it can fit the
data. Hence such $r$ has smaller empirical loss, but is not necessarily better
since it has higher variance.
Clearly, too inflexible $r$ also lead to a bad fit (``high bias'').

%-------------------------------%
\paradot{Main goal}
%-------------------------------%
The main goal of the paper is to establish a selection criterion in order to specify the smallest model
to which $f_{{\text{true}}}$ belongs or is close to, and simultaneously determine
the ``best'' fitting function $r(D)$. The criterion
\vspace{-1ex}\begin{itemize}\parskip=0ex\parsep=0ex\itemsep=0ex
\item is based on $r$ given as a black box that does not require insight into the
origin or inner structure of $r$;
\item does not depend on things often not given (like a stochastic noise model); and
\item exploits what is or should be given (like the loss function).
\vspace{-1ex}\end{itemize}
%

%-------------------------------%
\paradot{Definition of loss rank}
%-------------------------------%
We first consider discrete $\Y$ (i.e., classification), fix $\v x$, $\v y$ is
the observed data and $\v y'$ are fictitious others.
The key observation we exploit is that a more flexible $r$ can fit
more data $D'\in\D$ well than a more rigid one.
The more flexible $r$ is, the smaller the empirical loss $\L_r(\v y|\v x)$ is.
Instead of minimizing the unsuitable $\L_r(\v y|\v x)$ w.r.t.\ $r$,
we could ask how many $\v y'\in\Y^n$ lead to smaller $\L_r$ than $\v y$.
We define the loss rank of $r$ (w.r.t. $\v y$) as the number of $\v y'\in\Y^n$ with
smaller or equal empirical loss than $\v y$:
\beq\label{eqRank}
  \Rank_r(\v y|\v x) \equiv \Rank_r(L)
  := \#\{\v y'\!\in\!\Y^n : \L_r(\v y'|\v x)\!\leq\!L\}
  \;\;\mbox{with}\;\; L := \L_r(\v y|\v x)
\eeq
We claim that the loss rank of $r$ is a suitable model selection measure.
For \eqref{eqRank} to make sense, we have to assume (and will later assure)
that $\Rank_r(L)<\infty$, i.e., there are only
finitely many $\v y'\in\Y^n$ having loss smaller than $L$.

Since the logarithm is a strictly monotone increasing function, we
can also consider the logarithmic rank $\LR_r(\v y|\v
x):=\log\Rank_r(\v y|\v x)$, which will be more convenient.

%-------------------------------%
\begin{principle}[LoRP for classification]\label{prLRD}
%-------------------------------%
For discrete $\Y$, the best classifier/regressor $r:\D\times\X\to\Y$
in some class $\R$ for data $D=(\v x,\v y)$ is the one with the smallest
loss rank:
\beq\label{eqLRD}
  r^{best} \;=\; \arg\min_{r\in\R} \LR_r(\v y|\v x)
  \;\equiv\; \arg\min_{r\in\R} \Rank_r(\v y|\v x)
\eeq
where $\Rank_r$ is defined in \eqref{eqRank}.
\end{principle}

We give a simple example for which we can compute all ranks by hand
to help the reader better grasp how the principle works.

%-------------------------------%
\fexample{exSD}{simple discrete}{
%-------------------------------%
Consider $\X=\{1,2\}$, $\Y=\{0,1,2\}$, and two points
$D=\{(1,1),(2,2)\}$ lying on the diagonal $x=y$, with polynomial
(zero, constant, linear) least squares regressors
$\R=\{r_0,r_1,r_2\}$ (see Ex.\ref{exPR}). $r_0$ is simply 0, $r_1$
the $y$-average, and $r_2$ the line through points $(1,y_1)$ and
$(2,y_2)$. This, together with the quadratic $\L$ for generic $\v
y'$ and observed $\v y=(1,2)$ and fixed $\v x=(1,2)$, is
summarized in the following table
\beqn
\begin{array}{c|c|c|c}
  d & r_d(x|\v x,\v y') & \L_d(\v y'|\v x) & \L_d(D) \\ \hline
  0 &       0        & y'_1\!\,^2+y'_2\!\,^2 & 5         \\
  1 & \fr12(y'_1+y'_2)  & \fr12(y'_2-y'_1)^2  & \fr12       \\
  2 & (y'_2-y'_1)(x-1)+y'_1 & 0             & 0
\end{array}
\eeqn
From the $\L$ we can easily compute the Rank for all nine $\v
y'\in\{0,1,2\}^2$. Equal rank due to equal loss is
indicated by a ``$=$" in the table below. Whole equality groups are
actually assigned the rank of their right-most member, e.g.,\ for
$d=1$ the ranks of $(y'_1,y'_2)=(0,1),(1,0),(2,1),(1,2)$ are all 7 (and not
4,5,6,7).
\beqn\def\trq{\hspace{4.2ex}}
\begin{array}{c|c|c}
    & \Rank_{r_d}(y'_1y'_2|12) \\
  d & \quad\trq 1 \trq 2 \trq 3 \trq 4 \trq 5 \trq 6 \trq 7 \trq 8 \trq 9 \!\! & \Rank_{r_d}(D)\\ \hline
  0 & y'_1 y'_2 = 00<01=10<11<02=20<21={\bf 12}<22                             & 8             \\
  1 & y'_1 y'_2 = 00=11=22<01=10=21={\bf 12}<02=20                             & 7             \\
  2 & y'_1 y'_2 = 00=01=02=10=11=20=21=22={\bf 12}                             & 9             \\
\end{array}
\eeqn
So LoRP selects $r_1$ as best regressor, since it has minimal rank
on $D$. $r_0$ fits $D$ too badly and $r_2$ is too flexible (perfectly
fits all $D'$).
} % End of Example

%-------------------------------%
\paradot{LoRP for continuous $\Y$}
%-------------------------------%
We now consider the case of continuous or measurable spaces $\Y$,
i.e., normal regression problems. We assume $\Y=\SetR$ in the
following exposition, but the idea and resulting principle hold for
more general measurable spaces like $\SetR^m$. We simply reduce the
model selection problem to the discrete case by considering the
discretized space $\Y_\eps=\eps\SetZ$ for small $\eps>0$ and
discretize $\v y\leadsto \v y_\eps\in\eps\SetZ^n$ (``$\leadsto$" means ``is replaced by"). Then
$\Rank_r^\eps(L):=\#\{\v y'_\eps\in\Y_\eps^n : \L_r(\v y'_\eps|\v
x)\leq L\}$ with $L=\L_r(\v y_\eps|\v x)$ counting the number of
$\eps$-grid points in the set
\beq\label{defV}
  V_r(L) \;:=\; \{\v y'\in\Y^n : \L_r(\v y'|\v x)\leq L\}
\eeq
which we assume (and later assure) to be finite, analogous to the
discrete case. Hence $\Rank_r^\eps(L)\cdot\eps^n$ is an
approximation of the {\em loss volume} $|V_r(L)|$ of set $V_r(L)$, and
typically
$\Rank_r^\eps(L)\cdot\eps^n=|V_r(L)|\cdot(1+O(\eps))\to|V_r(L)|$ for
$\eps\to 0$. Taking the logarithm we get $\LR_r^\eps(\v y|\v
x)=\log\Rank_r^\eps(L)=\log|V_r(L)|-n\log\eps +O(\eps)$. Since
$n\log\eps$ is independent of $r$, we can drop it in comparisons
like \eqref{eqLRD}. So for $\eps\to 0$ we can define the log-loss
``rank'' simply as the log-volume
\beq\label{defLRC}
  \LR_r(\v y|\v x) \;:=\; \log|V_r(L)|,
  \qmbox{where} L:=\L_r(\v y|\v x)
\eeq

%-------------------------------%
\begin{principle}[LoRP for regression]\label{prLRC}
%-------------------------------%
For measurable $\Y$, the best regressor $r:\D\times\X\to\Y$ in some
class $\R$ for data $D=(\v x,\v y)$ is the one with the smallest loss
volume:
\beqn\label{eqLRC}
  r^{best} \;=\; \arg\min_{r\in\R} \LR_r(\v y|\v x)
  \;\equiv\; \arg\min_{r\in\R} |V_r(L)|
\eeqn
where $\LR$, $V_r$, and $L$ are defined in \eqref{defV} and \eqref{defLRC},
and $|V_r(L)|$ is the volume of $V_r(L)\subseteq\Y^n$.
\end{principle}

For discrete $\Y$ with counting measure we recover the discrete LoRP (Principle \ref{prLRD}).

%-------------------------------%
\fexample{exSC}{simple continuous}{
%-------------------------------%
Consider Example \ref{exSD} but with interval $\Y=[0,2]$.
The first table remains unchanged, while the second table becomes
\beqn
\begin{array}{c|c|c|c|c}
  d & V_d(L)=\{\v y'\in[0,2]^2: ...\} & |V_d(L)| &\L_d(D)& |V_d(\L_d(D))| \\ \hline\hline
  0 & y'_1\!\,^2+y'_2\!\,^2\leq L  & { \fr\pi4 L \ \text{if} \ L\leq4; \ \ \ \ \
    4 \ \text{if} \ L\geq 8; \atop \!\!2\sqrt{L-4}+L(\fr\pi4-\cos^{-1}(\fr{2}{\sqrt{L}})) \ \text{else}\!\!\!}
      & 5 & \doteq 3.6 \\ \hline
  1 & \fr12(y'_2-y'_1)^2\leq L      & { 4\sqrt{2L}-2L \ \text{if} \ L\leq 2; \atop 4 \ \text{if} \ L\geq2 } &\fr12& 3 \\ \hline
  2 & 0\leq L                      & 4 &0& 4
\end{array}
\eeqn
So LoRP again selects $r_1$ as best regressor, since it has smallest loss volume
on $D$.
} % End of Example

%-------------------------------%
\paradot{Infinite rank or volume}
%-------------------------------%
Often the loss rank/volume will be infinite, e.g.,\ if we had chosen
$\Y=\SetZ$ in Ex.\ref{exSD} or $\Y=\SetR$ in Ex.\ref{exSC}.
There are various potential remedies. We could
modify (a) the regressor $r$ or %
(b) the $\L$ to make $\LR_r$ finite, %
(c) the Loss Rank Principle itself, or %
(d) find problem-specific solutions. %
Regressors $r$ with infinite rank might be rejected for
philosophical or pragmatic reasons. We will briefly consider (a) for
linear regression later, but to fiddle around with $r$ in a generic
(blackbox way) seems difficult. We have no good idea how to tinker
with LoRP (c), and also a patched LoRP may be less attractive. For
kNN on a grid we later use remedy (d). While in (decision) theory,
the application's goal determines the loss, in practice the loss is
often more determined by convenience or rules of thumb. So the $\L$
(b) seems the most inviting place to tinker with. A very simple
modification is to add a small penalty term to the loss.
\beq\label{eqLa}
   \L_r(\v y|\v x) \;\leadsto\;
   \L_r^\a(\v y|\v x) := \L_r(\v y|\v x)+\a\|\v y\|^2,
   \quad \a>0 \mbox{ ``small''}
\eeq
The Euclidean norm $\|\v y\|^2:=\sum_{i=1}^n y_i^2$ is default, but
other (non)norm regularizations are possible. The regularized
$\LR_r^\a(\v y|\v x)$ based on $\L_r^\a$ is always finite, since
$\{\v y:\|\v y\|^2\leq L\}$ has finite volume.
An alternative penalty $\a\v{\h y}^\trp\v{\h y}$, quadratic in
the regression estimates $\h y_i=r(x_i|\v x,\v y)$ is possible if
$r$ is unbounded in every $\v y\to\infty$ direction.

A scheme trying to determine a single (flexibility) parameter (like $d$
and $k$ in the above examples) would be of no use if it depended
on one (or more) other unknown parameters ($\a$), since varying through
the unknown parameter leads to any (non)desired result.
Since LoRP seeks the $r$ of smallest rank, it is natural to also
determine $\a=\a_{\min}$ by minimizing $\LR_r^\a$ w.r.t.\ $\a$. The good news
is that this leads to meaningful results.
Interestingly, as we will see later, a clever choice of $\a$
may also result in alternative optimalities of the selection procedure.

%%%%%%%%%%%%%%%%%%%%%%%%%%%%%%%%%%%%%%%%%%%%%%%%%%%%%%%%%%%%%%%
\section{LoRP for y-Linear Models}\label{secLM}
%%%%%%%%%%%%%%%%%%%%%%%%%%%%%%%%%%%%%%%%%%%%%%%%%%%%%%%%%%%%%%%

In this section we consider the important class of {\it y-linear} regressions
with quadratic loss function.
By ``y-linear regression", we mean the linearity is only assumed in $y$
and the dependence on $x$ can be arbitrary. This class is richer
than it may appear. It includes the normal linear regression model, kNN (Example \ref{exKNN2}), kernel
(Example \ref{exKern}), and many other regression models. For y-linear
regression and $\Y=\SetR$, the loss rank is the volume of an
$n$-dimensional ellipsoid, which can efficiently be computed in time
$O(n^3)$ (Theorem \ref{thmLRL}). For the special case of projective
regression, e.g.,\ linear basis function regression (Example
\ref{exLBFR}), we can even determine the regularization parameter
$\a$ analytically (Theorem \ref{thmLRP}).

%-------------------------------%
\paradot{y-Linear regression}
%-------------------------------%
We assume $\Y=\SetR$ in this section; generalization to $\SetR^m$ is
straightforward. A y-linear regressor $r$ can be written in the form
\beq\label{eqmj}
  \h y \;=\; r(x|\v x,\v y) \;=\; \sum_{j=1}^n m_j(x,\v x)y_j
   \quad\forall x\in\X
   \qmbox{and some} m_j:\X\times\X^n\to\SetR
\eeq
Particularly interesting is $r$ for $x=x_1,...,x_n$.
\beq\label{eqM}
  \h y_i \;=\; r(x_i|\v x,\v y) \;=\; \sum_j M_{ij}(\v x)y_j
  \qmbox{with} M:\X^n\to\SetR^{n\times n}
\eeq
where matrix $M_{ij}(\v x)=m_j(x_i,\v x)$. Since LoRP needs $r$ only on the
training data $\v x$, we only need $M$.

%-------------------------------%
\fexample{exKNN2}{kNN ctd.}{
%-------------------------------%
For kNN of Ex.\ref{exKNN} we have
$m_j(x,\v x)={1\over k}$ if $j\in\N_k(x)$ and 0 else, and
$M_{ij}(\v x)={1\over k}$ if $j\in\N_k(x_i)$ and 0 else.
} % End of Example

%-------------------------------%
\fexample{exKern}{kernel regression}{
%-------------------------------%
Kernel regression takes a weighted average over $\v y$,
where the weight of $y_j$ to $y$ is proportional to
the similarity of $x_j$ to $x$, measured by a kernel
$K(x,x_j)$, i.e., $m_j(x,\v x)=K(x,x_j)/\sum_{j=1}^n K(x,x_j)$.
For example the Gaussian kernel for $\X=\SetR^m$ is
$K(x,x_j)=\e^{-\|x-x_j\|_2^2/2\s^2}$.
The width $\s$ controls the smoothness of the kernel regressor,
and LoRP selects the real-valued ``complexity'' parameter $\s$.
} % End of Example

%-------------------------------%
\fexample{exLBFR}{linear basis function regression, LBFR}{
%-------------------------------%
Let $\phi_1(x),...,\phi_d(x)$ be a set or vector of ``basis''
functions often called ``features''. We place no restrictions on
$\X$ or $\v\phi:\X\to\SetR^d$. Consider the class of functions
linear in $\v\phi$:
\beqn
  \F_d \;=\; \{ f_{\v w}(x)=\textstyle{\sum_{a=1}^d} w_a\phi_a(x)=\v w^\trp\v\phi(x) : \v w\in\SetR^d \}
\eeqn
For instance, for $\X=\SetR$ and $\phi_a(x)=x^{a-1}$ we would recover
the polynomial regression Example \ref{exPR}.
For quadratic loss function $\L(y_i,\h y_i)=(y_i-\h y_i)^2$ we have
\beqn
  \L_{\v w}(\v y|\v\phi)
  \;:=\; \sum_{i=1}^n(y_i-f_{\v w}(x_i))^2
  \;=\; \v y^\trp\v y - 2\v y^\trp\Phi\v w + \v w^\trp B \v w
\eeqn
where matrix $\Phi$ is defined by $\Phi_{ia}=\phi_a(x_i)$ and
$B$ is a symmetric matrix with
$B_{ab}=\sum_{i=1}^n\phi_a(x_i)\phi_b(x_i)=[\Phi^\trp\Phi]_{ab}$.
The loss is quadratic in $\v w$ with minimum at $\v
w=B^{-1}\Phi^\trp\v y$. So the least squares regressor is $\h y=\v
y^\trp\Phi B^{-1}\v\phi(x)$, hence $m_j(x,\v x)=(\Phi
B^{-1}\v\phi(x))_j$ and $M(\v x)=\Phi B^{-1}\Phi^\trp$.
} % End of Example

Consider now a general linear regressor $M$ with quadratic loss
and quadratic penalty
\vspace{-3ex}\bqa\nonumber
  \L_M^\a(\v y|\v x) &=& \sum_{i=1}^n
  \left(y_i-\textstyle{\sum_{j=1}^n} M_{ij}y_j\right)^2+\a\|\v y\|^2
  \;=\; \v y^\trp S_\a\v y,
\\ \label{defSa}
  \qmbox{where\footnotemark} S_\a &=& (I-M)^\trp(I-M)+\a I
\eqa%
\footnotetext{The mentioned alternative penalty $\a\|\v{\h
y}\|^2$ would lead to
$S_\a = (I-M)^\trp(I-M)+\a M^\trp M$.
For LBFR, penalty $\a\|\v{\h w}\|^2$ is popular (ridge regression).
Apart from being limited to parametric regression, it
has the disadvantage of not being reparametrization invariant.
For instance, scaling $\phi_a(x)\leadsto\g_a\phi_a(x)$ does not
change the class $\F_d$, but changes the ridge regressor.}
($I$ is the identity matrix). $S_\a$ is a symmetric matrix. For
$\a>0$ it is positive definite and for $\a=0$ positive semidefinite.
If $\l_1,...,\l_n\geq 0$ are the eigenvalues of $S_0$, then
$\l_i+\a$ are the eigenvalues of $S_\a$. $V(L)=\{\v y'\in\SetR^n : \v
y'\!\,^\trp S_\a\v y'\leq L\}$ is an ellipsoid with the eigenvectors of
$S_\a$ being the main axes and $\sqrt{L/(\l_i+\a)}$ being their length.
Hence the volume is
\beqn
  |V(L)| \;=\; v_n\prod_{i=1}^n \sqrt{L\over \l_i+\a}
  \;=\; {v_n L^{n/2}\over\sqrt{\det S_\a}}
\eeqn
where $v_n=\pi^{n/2}/{n\over 2}!$ is the volume of the
$n$-dimensional unit sphere, $z!:=\G(z+1)$, and $\det$ is the
determinant. Taking the logarithm we get
\beq\label{eqLRL1}
  \LR_M^\a(\v y|\v x)
  \;=\; \log|V(\L_M^\a(\v y|\v x))|
  \;=\; \fr n2\log(\v y^\trp S_\a\v y)-\fr12\log\det S_\a +\log v_n
\eeq
Since $v_n$ is independent of $\a$ and $M$ it is possible to drop $v_n$.
Consider now a class of linear regressors $\M=\{M\}$,
e.g.,\ the kNN regressors $\{M_k:k\in\SetN\}$ or
the $d$-dimensional linear basis function regressors $\{M_d:d\in\SetN_0\}$.

%-------------------------------%
\begin{theorem}[LoRP for y-linear regression]\label{thmLRL}
%-------------------------------%
For $\Y=\SetR$, the best linear regressor $M:\X^n\to\SetR^{n\times n}$
in some class $\M$ for data
$D=(\v x,\v y)$ is
\beq\label{eqLRL}
  M^{best} \;=\; \mathop{\arg\min}_{M\in\M,\a\geq 0}
  \{ \fr n2\log(\v y^\trp S_\a\v y)-\fr12\log\det S_\a \}
  \;=\; \mathop{\arg\min}_{M\in\M\;\a\geq 0}
  \Big\{ {\v y^\trp S_\a\v y\over(\det S_\a)^{1/n}} \Big\}
\eeq
where $S_\a=S_\a(M)$ is defined in \eqref{defSa}.
\end{theorem}

The last expression shows that linear LoRP minimizes the $\L$
times the geometric average of the squared axes lengths of ellipsoid
$V(1)$. Note that $M^{best}$ depends on $\v y$ unlike the $M\in\M$.

%-------------------------------%
\paradot{Nullspace of $S_0$}
%-------------------------------%
If $M$ has an eigenvalue 1, then $S_0=(I-M)^\trp(I-M)$ has a zero
eigenvalue and $\a>0$ is necessary, since $\det S_0=0$. Actually
this is true for most practical $M$. Most linear regressors
are invariant under a constant shift of $\v y$, i.e., $r(x|\v
x,\v y+c)=r(x|\v x,\v y)+c$, which implies that $M$ has
eigenvector $(1,...,1)^\trp$ with eigenvalue 1. This can easily be
checked for kNN (Ex.\ref{exKNN}), kernel (Ex.\ref{exKern}), and LBFR
(Ex.\ref{exLBFR}). Such a generic 1-eigenvector effecting all
$M\in\M$ could easily and maybe should be filtered out by
considering only the orthogonal space or dropping these $\l_i=0$
when computing $\det S_0$. The 1-eigenvectors that depend on $M$ are
the ones where we really need a regularizer $\a>0$. For instance,
$M_d$ in LBFR has $d$ eigenvalues 1, and $M_{\text{kNN}}$ has as
many eigenvalues 1 as there are disjoint components in the graph
determined by the edges $M_{ij}>0$.
In general we need to find the optimal $\a$ numerically.
If $M$ is a projection we can find $\a_m$ analytically.

%------------------------------------------------%
\paradot{Numerical approximation of $(\det S_{\a})^{1/n}$ and the computational complexity of linear LoRP}
%------------------------------------------------%
For each $\a$ and candidate model,
the determinant of $S_\a$ in the general case can be computed in time $O(n^3)$.
Often $M$ is a very sparse matrix (like in kNN) or can be well approximated by a sparse matrix
(like for kernel regression), which allows us to approximate $\det S_\a$ sometimes in linear
time \citep{Reusken:02}.
To search the optimal $\a$ and $M$, the computational cost depends on
the range of $\a$ we search and the number of candidate models we have.

%-------------------------------%
\paradot{Projective regression}
%-------------------------------%
Consider a projection matrix $M=P=P^2$ with $d(=\tr P)$ eigenvalues 1,
and $n-d$ zero eigenvalues.
For instance, $M=\Phi B^{-1}\Phi^\trp$ of LBFR Ex.\ref{exLBFR} is such a matrix.
This implies
that $S_\a$ has $d$ eigenvalues $\a$ and $n-d$ eigenvalues $1+\a$,
thus $\det S_\a = \a^d(1+\a)^{n-d}$.
Let $\r=\|\v y-\hat{\v y}\|^2/\|\v y\|^2$,
then $\v y^\trp S_\a\v y = (\r+\a)\v y^\trp\v y$ and
\beq\label{eqLRPa}
\LR_P^\a = \fr n2\log\v y^\trp\v y + \fr n2\log(\r+\a)-
  {\textstyle{d\over 2}}\log\a - {\textstyle{n-d\over 2}}\log(1+\a).
\eeq
Solving $\partial\LR_P^\a/\partial\a = 0$ w.r.t.\ $\a$ we get a minimum at
$\a=\a_m:={\r d\over(1-\r)n-d}$ provided that $1-\r>{d/ n}$.
After some algebra we get
\beq\label{eqLRPamin}\textstyle
  \LR_P^{\a_m} = \fr n2\log\v y^\trp\v y - \fr n2 \KL({d\over n}\|1-\r),
  \qmbox{where} \KL(p\|q): = p\log{p\over q}+(1-p)\log{1-p\over 1-q}
\eeq
is the relative entropy or the Kullback-Leibler divergence.
Note that \eqref{eqLRPamin} is still valid without the condition $1-\rho>{d}/{n}$
(the term $\log((1-\rho)n-d)$ has been canceled in the derivation).
What we need when using \eqref{eqLRPamin} is that $d<n$ and $\rho<1$,
which are very reasonable in practice.
Interestingly, if we use the penalty $\a\|\hat{\v y}\|^2$ instead of $\a\|\v y\|^2$,
the loss rank then has the same expression as \eqref{eqLRPamin} without any condition\footnote{
Then $S_\a=(I_n- P)^\trp(I_n-P)+\a P^\trp P=I_n+(\a-1) P$
has $d$ eigenvalues $\a$ and $n-d$ eigenvalues 1, thus $\det(S_\a)=\a^{d}$.
The loss rank $\LR_P^\a=\fr n2\log\v y^\trp\v y+\fr n2\log(1+(\a-1)(1-\rho))-\fr{d}{2}\log\a$
is minimized at $\a_m=\fr{\rho d}{(1-\rho)(n-d)}$.
After some algebra we get the same expression of $\LR_P^{\a_m}$ as \eqref{eqLRPamin}.}.

Minimizing $\LR_P^{\a_m}$ w.r.t.\ $P$ is equivalent to
maximizing $\KL({d\over n}\|1-\r)$.
The term $\r$ is a measure of fit.
If $d$ increases, then $\rho$ decreases and otherwise.
We are seeking a tradeoff between the model complexity $d$ and the measure of fit $\r$,
and LoRP suggests the optimal tradeoff by maximizing $\KL$.

%-------------------------------%
\begin{theorem}[LoRP for projective regression]\label{thmLRP}
%-------------------------------%
The best projective regressor $P:\X^n\to\SetR^{n\times n}$
with $P=P^2$ in some projective class $\cal P$ for data
$D=(\v x,\v y)$ is
\beq\label{eqLRP}
  P^{best} \;=\; \arg\max_{P\in\cal P} \;\textstyle \KL({\tr P(\v x)\over n}\|{\v y^\trp P(\v x)\v y\over\v y^\trp\v y}).
\eeq
\end{theorem}

%%%%%%%%%%%%%%%%%%%%%%%%%%%%%%%%%%%%%%%%%%%%%%%%%%%%%%%%%%%%%%%
\section{Optimality Properties of LoRP for Variable Selection}\label{secOpt}
%%%%%%%%%%%%%%%%%%%%%%%%%%%%%%%%%%%%%%%%%%%%%%%%%%%%%%%%%%%%%%%

In the previous sections, LoRP was stated for general-purpose model selection.
By restricting attention to linear regression models, we will point
out in this section some theoretical properties of LoRP for variable
(also called feature or attribute) selection.

Variable selection is probably the most fundamental and important
topic in linear regression analysis. At the initial stage of
modeling, a large number of potential covariates are often
introduced; one then has to select a smaller subset of the
covariates to fit/interpret the data. There are two main goals of
variable selection, one is model identification, the other is
regression estimation. The former aims at identifying the true
subset generating the data, while the latter aims at estimating
efficiently the regression function, i.e., selecting a subset that
has the minimum mean squared error loss. Note that whether or not
there is a selection criterion achieving simultaneously these two
goals is still an open question \citep{Yang:05, Gruenwald:04}. We
show that with the optimal parameter $\a$ (defined as $\a_m$ that
minimizes the loss rank $\LR_M^\a$ in $\a$), LoRP satisfies the
first goal, while with a suitable choice of $\a$, LoRP satisfies the
second goal.

Given $d+1$ potential covariates $X_0\equiv1,X_1,...,X_d$ and
a response variable $Y$, let $ X=\v x$ be a non-random design matrix
of size $n\times(d+1)$ and $\v y$ be a response vector respectively
(if $\v y$ and $X$ are centered, then the covariate 1 can be omitted
from the models). Denote by $\S=\{0,j_1,...j_{|\S|-1}\}$ the candidate
model that has covariates $X_0,X_{j_1},...,X_{j_{|\S|-1}}$. Under a
proposed model $\S$, we can write
\beqn
  \v y \;=\; X_\S\v\beta_\S+\s_\S\v\epsilon
\eeqn
where $\epsilon$ is noise with expectation $\E[\v\epsilon]=0$ and
covariance $\text{Cov}(\v\epsilon)=I_n$, $\s_\S>0$,
$\v\beta_\S=(\beta_0,\beta_{j_1},...,\beta_{j_{|\S|-1}})^\trp$, and
$ X_\S$ is the $n\times |\S|$ design matrix obtained from $ X$ by
removing the $(j+1)$st column for all $j\not\in\S$.

%-------------------------------------------------%
\paradot{Model consistency of LoRP for variable selection}
%-------------------------------------------------%
The ordinary least squares (OLS) fitted vector under model $\S$ is
$\hat{\v y}_\S= M_\S\v y\;\;\text{with}\;\;  M_\S= X_\S( X_\S^\trp X_\S)^{-1} X_\S^\trp$
being a projection matrix.
From Theorem \ref{thmLRP} the best subset chosen by LoRP is
\beqn
  \hat{\S}_n \;=\; \arg\min_\S\LR_\S^{\a_m} \;=\; \arg\max_\S\{\KL(\fr{|\S|}{n}\|1-\rho_\S)\},\;\;\rho_\S=\fr{\|\v y-\hat{\v y}_\S\|^2}{\|\v y\|^2}.
\eeqn
The term $\rho_\S$ is a measure of fit. It will be very close to 0 if model $\S$ is big,
otherwise, it will be close to 1 if $\S$ is too small.
Therefore, it is reasonable to consider only cases in which $\rho_\S$ is bounded away from $0$ and $1$.
In order to prove the theoretical properties of LoRP, we need the following technical assumption.
\begin{itemize}
\item[(A)] For each candidate model $\S$, $\rho_\S$ is bounded away from 0 and 1, i.e., there are constants $c_1$ and $c_2$ such that
           $ 0<c_1\leq \rho_\S\leq c_2<1$ with probability 1 (w.p.1).
\end{itemize}
Let $\hat\s_\S^2=\|\v y-\hat{\v y}_\S\|^2/n$ and $\S_\text{null}=\{0\}$.
It is easy to see that for every $\S$
\beq\label{srho}
  1-\rho_\S = \|\hat{\v y}_\S\|^2 / \|\v y\|^2,\qquad
  n\hat\s_\S^2 = \rho_\S \|\v y\|^2, \qquad
  n\,\bar{\v y}^2 = \|\hat{\v y}_{\S_\text{null}}\|^2
  \leq \|\hat{\v y}_\S\|^2 \leq \|\v y\|^2
\eeq
where $\bar{\v y}$ denotes the arithmetic mean $\sum_{i=1}^n y_i/n$.
Assumption (A) follows from
\begin{itemize}
\item[(A')] $0<\liminf\limits_{n\to\infty}(\bar{ \v y})^2\leq\limsup\limits_{n\to\infty}(\fr1n\|{\v y}\|^2)<\infty$ \ \ and \ \ $\forall\S:$ $\hat\s_\S^2\to\s_\S^2>0$ w.p.1.
\end{itemize}
The first condition of (A') is obviously very mild
and satisfied in almost all cases in practice.
The second one is routinely used to derive
asymptotic properties of model selection criteria
(e.g., Theorem 2 of \cite{Shao:97} and Condition 1 of \cite{Wang:07}).

%-------------------------------------------------%
\begin{lemma}[LoRP for variable selection]\label{LemmaVariableSelection}
%-------------------------------------------------%
The loss rank of model $\S$ is
\beq\label{varsel}
  \LR_\S \;\equiv\; \LR_\S^{\a_m} \;=\;
  \fr n2\log(n\hat\s_\S^2)+\fr n2 H(\fr{|\S|}{n}) +\fr d2\log\fr{1-\rho_\S}{\rho_\S}
\eeq
where $\rho_\S$ and $\hat\s_\S^2$ are defined in \eqref{srho}, and
$H(p):=-p\log p-(1-p)\log(1-p)$ is the entropy of $p$. Under
Assumption (A) or (A'), after neglecting constants independent of
$\S$, the loss rank of model $\S$ has the form
\beq\label{BICtype}
  \LR_\S \;=\; \fr n2\log\hat\s_\S^2+\fr{|\S|}{2} \log n +O_\P(1),
\eeq
where $O_\P(1)$ denotes a bounded random variable w.p.1.
\end{lemma}

%-------------------------------------------------%
\paradot{Proof}
%-------------------------------------------------%
Inserting $\v y^\trp\v y=n\hat\s_\S^2/\rho_\S$ into \eqref{eqLRPamin}
and rearranging terms gives \eqref{varsel}. By Assumption (A) the last
term in \eqref{varsel} is bounded w.p.1. Taylor expansion
$\log(1-p)=-p+O(p^2)$ implies $H(p)/p+\log p\to 1$, hence $\fr n2
H(\fr{|\S|}{n})=\fr{|\S|}{2}\log n+O(1)$. Finally, dropping the
$\S$-independent term $\fr n2\log n$ from \eqref{varsel} gives
\eqref{BICtype}.
\qed

This lemma implies that the loss rank $\LR_\S$ here is a BIC-type criterion,
thus we immediately can state without proof the following theorem
which is the well-known model consistency of BIC-type criteria
(interested readers can find the routine proof in, for example, \cite{Cha:06}).

%----------------------------------------------%
\begin{theorem}[Model consistency]\label{consistency}
%---------------------------------------------%
Under Assumption (A) or (A'), LoRP is model consistent for variable
selection in the sense that the probability of selecting the true
model goes to 1 for data size $n\to\infty$.
\end{theorem}

%-------------------------------%
\paradot{The optimal regression estimation of LoRP}
%-------------------------------%
The second goal of model selection is
often measured by the (asymptotic) mean efficiency \citep{Shibata:83} which is briefly defined as follows.
Let $\S_T$ denote the true model (which may contain an infinite number of covariates).
For a candidate model $\S$, let
$L_n(\S)=\| X_{\S_T}\v \beta_{\S_T}- X_\S\hat{\v\beta}_\S\|^2$
be the squared loss where $\hat{\v\beta}_\S$ is the OLS estimate,
and $R_n(\S)=\E[L_n(\S)]$ be the risk.
The mean efficiency of a selection criterion $\delta$ is defined by the ratio
\beqn
\text{eff}(\delta)=\dfrac{\inf_\S R_n(\S)}{\E[L_n(\S_\delta)]}\leq 1
\eeqn
where $\S_\delta$ is the model selected by $\delta$.
$\delta$ is said to be asymptotically mean efficient if $\liminf_{n\to\infty}\text{eff}(\delta)=1$.

By minimizing the loss rank in $\a$ we have shown in the previous paragraph
that LoRP satisfies the first goal of model selection.
We now show that with a suitable choice of $\a$,
LoRP also satisfies the second goal.

From \eqref{eqLRPa}, we have
\beqn
  \LR_\S^\a(\v y|\v x) = \fr n2\log(\hat\s^2_\S+\fr{\a}{n}\v y^\trp \v y)+\fr n2\log n-\fr{|\S|}2\log(\a)-\fr{n-|\S|}{2}\log(1+\a).
\eeqn
By choosing $\a=\tilde\a=\exp(-\fr{n(n+|\S|)}{|\S|(n-|\S|-2)})$,
under Assumption (A),
the loss rank of model $\S$ (neglecting the common constant $\fr n2\log n$) is proportional to
\beqn
  \LR_\S^{\tilde\a}(\v y|\v x)=n\log{\hat\s}^2_\S+\fr{n(n+|\S|)}{n-|\S|-2}+o_{\P}(1),
\eeqn
which is the corrected AIC of \cite{Hurvich:89}. As a result,
LoRP$(\tilde\a)$ is optimal in terms of regression estimation, i.e.,
it is asymptotically mean efficient (\citeauthor{Shibata:83}, 1983;
\citeauthor{Shao:97}, 1997).

%------------------------------------------------%
\begin{theorem}[Asymptotic mean efficiency]\label{efficiency}
Under Assumption (A) or (A'), with a suitable choice of $\a$,
the loss rank is proportional to the corrected AIC.
As a result, LoRP is asymptotically mean efficient.
\end{theorem}

%%%%%%%%%%%%%%%%%%%%%%%%%%%%%%%%%%%%%%%%%%%%%%%%%%%%%%%%%%%%%%%
\section{Experiments}\label{secExp}
%%%%%%%%%%%%%%%%%%%%%%%%%%%%%%%%%%%%%%%%%%%%%%%%%%%%%%%%%%%%%%%
In this section we present a simulation study for LoRP,
compare it to other methods
and also demonstrate how LoRP can be used for some specific problems
like choosing tuning parameters for kNN and spline regression.
All experiments are conducted by using MATLAB software and the source code is
freely available at \underline{http://www.hutter1.net/ai/lorpcode.zip}.

%-------------------------------%
\paradot{Comparison to AIC and BIC for model identification}
%-------------------------------%
Samples are generated from the model
\beq\label{model1}
y=\beta_0+\beta_1X_1+...+\beta_dX_d+\epsilon,\;\;\epsilon\sim N(0,\s^2)
\eeq
where $\v\beta$ is the vector of coefficients
with some zero entries.
Without loss of generality, we assume that $\beta_0=0$,
otherwise, we can center the response vector $\v y$ and standardize the design matrix $X$
to exclude $\beta_0$ from the model.
We shall compare the performance of LoRP
to that of BIC and AIC with various factors $n,\ d$ and signal-to-noise ratio (SNR) which is $\|\v\beta\|^2/\s^2$
($\|\v\beta\|^2$ is often called the length of the signal).

For a given set of factors $(n,\ d,\ \SNR)$, the way we simulate a dataset from model \eqref{model1} is as follows.
Entries of $X$ are sampled from a uniform distribution on $[-1,1]$.
To generate $\v\beta$, we first create a vector
$\v u = (u_1,...,u_d)^\trp$ whose entries are sampled from a uniform distribution on $[-1,1]$.
The number of true covariates $d^*$ is randomly selected from $\{1,2,...,d\}$,
the last $d-d^*$ entries of $\v u$ are set to zero,
then coefficient vector $\v\beta$ is computed by
$\beta_i=\{\text{length of signal}\}*u_i/||\v u||$.
In our simulation, the length of signal was fixed to be $10$.
$n$ observation errors $\epsilon_1,...,\epsilon_n$ are sampled from
a normal distribution with mean 0 and variance $\s^2=||\v\beta||^2/\SNR$.
Finally, the response vector is computed by $\v y=X\v\beta+\v\epsilon$.
For each set of factors $(n,\ d,\ \SNR)$, 1000 datasets are simulated in the same manner
to assess the average performance of the methods.
For simplicity, a candidate model is specified by its order,
i.e., we search the best model among only $d$ models $\{1\},\{1,2\}...,\{1,2,...,d\}$.
For the general case, an efficient branch-and-bound algorithm \citep[Chp.3]{Miller:02}
can be used to exhaustively search for the best subsets.

Table \ref{Table1} presents percentages of correctly-fitted models with various factors $n$, $d$ and SNR.
As shown, LoRP outperforms the others.
The better performance of LoRP over BIC,
which is the most popular criterion for model identification,
is very encouraging.
This is probably because LoRP is a selection criterion with a data-dependent penalty.
This improvement needs a theoretical justification which we intend to do in the future.

\begin{table}[ht]
\caption{Percentage of correctly-fitted models over 1000 replications} % title of Table
\centering % used for centering table
\begin{tabular}{cccccc|cccccc}
\hline\hline %inserts double horizontal lines
$n$ & $d$   &   SNR & AIC & BIC & LoRP  & $n$ & $d$   &   SNR & AIC & BIC & LoRP\\ \hline
100 & 5 & 1 & 62  & 62  & 69  & 300 & 5 & 1 & 74  & 82  & 83\\
  &   & 5 & 85  & 85  & 86  &   &   & 5 & 78  & 90  & 91\\
  &   & 10  & 80  & 90  & 91  &   &   & 10  & 81  & 94  & 94\\
  & 10  & 1 & 52  & 42  & 54  &   & 10  & 1 & 63  & 67  & 71\\
  &   & 5 & 63  & 77  & 77  &   &   & 5 & 70  & 85  & 86\\
  &   & 10  & 68  & 84  & 85  &   &   & 10  & 74  & 90  & 90\\
  & 20  & 1 & 32  & 22  & 36  &   & 20  & 1 & 54  & 45  & 61\\
  &   & 5 & 55  & 63  & 65  &   &   & 5 & 64  & 79  & 80\\
  &   & 10  & 56  & 73  & 74  &   &   & 10  & 67  & 85  & 85\\

\hline %inserts single line
\end{tabular}
\label{Table1} % is used to refer this table in the text
\end{table}

%-------------------------------%
\paradot{Comparison to AIC and BIC for regression estimation}
%-------------------------------%
Consider the following model which is from \cite{Shibata:83}
\beq\label{modelShibata}
y=y(x)=\log\fr1{1-x}+\epsilon,\;\;\epsilon\sim N(0,\s^2),\;x\in[0,1).
\eeq
We approximate the true function by a Fourier series
and consider the problem of choosing a good order among models
\beqn
y=\beta_0+\sum_{l=1}^{k-1}\fr{\cos(\pi lx/\delta)}{l+1}\beta_{l}+\epsilon,\;\;k=1,...,K.
\eeqn
In the present context, a model in Section \ref{secOpt}
is completely specified by the order $K$ of the Fourier series.
Samples are created from \eqref{modelShibata} at the points $x_i=\delta\fr{i}{n+1}$, $i=1,...,n$.
As in \cite{Shibata:83}, we take $\delta=.99$, and $K=163$ with various $n$ and $\s$.
The performance is measured by the estimate of mean efficiency over 1000 replications.

Table \ref{Table2} represents the simulation results.
In general, LoRP (with $\a=\tilde\a$ as in Section \ref{secOpt}) outperforms the others,
except for cases with unrealistically high noise level.
For cases with high noise, mean efficiency of BIC is often larger than that of AIC and LoRP.
This was also shown in the simulation study of \cite{Shibata:83}, Table 1.
This phenomenon can be explained as follows.

The risk of model $k$ (the model specified by its order $k$) is $R_n(k)=\|(I-M_k)\v y_{\text{true}}\|^2+k\s^2$
where $M_k$ is the regression matrix under model $k$ and $\v y_{\text{true}}$ is the vector of true values $y(x_i)$.
When $\s\to\infty$, the ideal $k^\star=\arg\inf_k R_n(k)\to1$.
Because BIC penalizes the model complexity more strongly than AIC and LoRP do,
the order chosen by BIC is closer to $k^\star=1$ than the ones chosen by AIC and LoRP.
As a result, mean efficiency of BIC is larger than that of the others.

\begin{table}[ht]
\caption{Estimates of mean efficiency over 1000 replications} % title of Table
\centering % used for centering table
\begin{tabular}{ccccc|ccccc}
\hline\hline %inserts double horizontal lines
$n$ &   $\s$  & AIC & BIC & LoRP  & $n$ & $\s$  & AIC & BIC & LoRP\\ \hline
400 &.001&1.00&.98&.99& 600 &.001&1.00&.98&1.00\\
  &.01&.93&.68&.90&   &.01&.99&.67&.92\\
  &.05&.88&.67&.95&   &.05&.90&.66&.94\\
  &.1&.88&.67&.92&    &.1&.90&.67&.93\\
  &.5&.81&.66&.85&    &.5&.82&.66&.83\\
  &1&.79&.63&.82&     &1&.79&.65&.82\\
  &5&.67&.65&.70&     &5&.65&.67&.66\\
  &10&.54&.67&.59&    &10&.54&.59&.54\\
  &100&.31&.89&.33&   &100&.40&.90&.41\\
\hline %inserts single line
\end{tabular}
\label{Table2} % is used to refer this table in the text
\end{table}
%-------------------------------%
\paradot{LoRP for selecting a good number of neighbors in kNN}
%-------------------------------%
Let us now see how LoRP can be applied to select a good parameter $k$ in kNN regression.

\begin{figure*}
\centerline{\includegraphics[width=1.2\textwidth,height=0.6\textwidth]{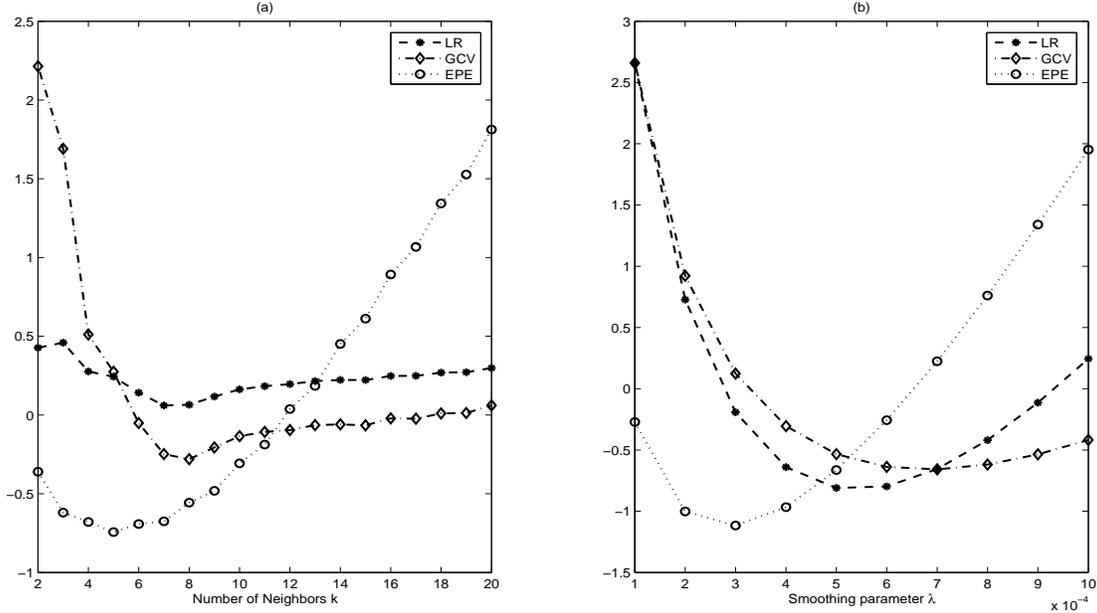}}
\caption{\label{figure}Choosing the tuning parameters in kNN and spline regression.
The curves have been scaled by their standard deviations.}
\end{figure*}

We created a dataset of $n=100$ observations $(x_i,y_i)$ from
the model:
\beq\label{equkNN}
  y=f(x)+\eps,\ \mbox{with}\ f(x)=\fr{\sin(12(x+0.2))}{x+0.2},\ x\in [0,1]
\eeq
where $\eps\sim N(0,\s^2)$ with $\s=0.5$. The regression matrix $M^{(k)}$
for kNN regression is determined by
$M_{ij}^{(k)}={1\over k}$ if $j\in\N_k(x_i)$ and 0 else. Then, the
loss rank is
\beqn
  \LR(k)=\inf_{\a\geq0}\{\fr{n}{2}\log(\v y^\trp S_{\a}^{(k)}\v y)-\fr12\log\det S_{\a}^{(k)}\},
\eeqn
where $S_{\a}^{(k)}=(I-M^{(k)})^\trp (I-M^{(k)})+\a
I$.
The most widely-used method to select a good $k$ is probably Generalized Cross-Validation (GCV) \citep{Craven:79}:
$\GCV(k)=n\|(I-M^{(k)})\v y\|^2/[\tr(I-M^{(k)})]^2$.
To judge how well GCV and LoRP work, we compare them to the expected prediction error defined as
\beqn
  \text{EPE}(k) \;=\; \sum_{i=1}^n \E(y_i-\hat y_i)^2
  \;=\; \sum_{i=1}^n\Big[\s^2+(f(x_i)-\fr1k\sum_{\nq j\in\N_k(x_i)\nq}f(x_j))^2+\fr{\s^2}{k}\Big].
\eeqn
Figure \ref{figure}(a) shows the
curves $\LR(k),\ \GCV(k),\ \text{EPE}(k)$ for $k=2, ...,20$
(the trivial case $k=1$ is omitted),
in which $k=7$-nearest neighbors is chosen by LoRP and $k=8$ is chosen by GCV.
The ``ideal" $k$ is 5.
Both LoRP and GCV do a reasonable job.
LoRP works slightly better than GCV.

%-------------------------------%
\paradot{LoRP for selecting a good smoothing parameter}
%-------------------------------%
We now further demonstrate the use of LoRP in selecting a good smoothing parameter for spline regression.
Consider the following problem: find a function belonging to the
class of functions with continuous 2nd derivative that minimizes the following
penalized residual sum of squares:
\beqn%\label{equsecSmosli1}
  \mbox{RSS}(f)=\sum_{i=1}^n(y_i-f(x_i))^2+\lambda\int (f''(t))^2dt,
\eeqn
where $\lambda$ is called the smoothing parameter. The second term
penalizes the curvature of function $f$ and the smoothing parameter
$\lambda$ controls the amount of penalty.
Our goal is to choose a good $\lambda$.

It is well known (see, e.g., \cite{Hastie:01}, Section 5.4) that the solution is a natural spline
$f(x)=\sum_{j=1}^nN_j(x)\theta_j$
where $N_1(x),. . . ,N_n(x)$ are the basis functions of the natural
cubic spline:
\beqn
  N_1(x)=1,\ N_2(x)=x,\ N_{k+2}(x)=d_k(x)-d_{n-1}(x)\;\;\text{with}\;\; d_k(x)=\fr{(x-x_k)_+^3-(x-x_n)_+^3}{x_n-x_k}.
\eeqn
The problem thus reduces to finding a vector $\v\theta\in\SetR^n$ that
minimizes
\beqn%\label{equsecSmosli3}
  \mbox{RSS}(\v\theta)=(\v y- N\v\theta)^\trp (\v y- N\v\theta)+\lambda\v\theta^\trp \Omega\v\theta
\eeqn
where $ N_{ij}=N_j(x_i)$ and $\Omega_{ij}=\int N_i''(x)N_j''(x)dx$. It is easy to see that the
solution is $ \v{\hat\theta}_\lambda=( N^\trp  N+\lambda\Omega)^{-1} N^\trp \v y$,
and the fitted vector is $\v {\hat y}= N\v{\hat\theta}_\lambda= M_\lambda\v y$ with
$ M_\lambda= N( N^\trp  N+\lambda\Omega)^{-1} N^\trp \v y$.
The fitted vector is linear in $\v y$, thus the loss rank is
\beqn%\label{equsecSmosli6}
\LR(\l)=\arg\min_{\alpha\ge0}\{\fr{n}{2}\log(\v y^\trp S_\lambda^\alpha\v y)-\fr12\log\det S_\lambda^\alpha\}
\eeqn
where $S_\lambda^\alpha=(I- M_\lambda)^\trp (I- M_\lambda)+\alpha I$.

Let us consider again the dataset generated from model \eqref{equkNN}.
Figure \ref{figure}(b) shows the curves $\LR(\lambda)$, $\GCV(\lambda)$ and $\text{EPE}(\lambda)$.
The derivation of expressions for $\GCV(\lambda)$ and $\text{EPE}(\lambda)$ is
similar to the previous example.
$\l\approx 3\times10^{-4}$ is the optimal value selected by the ``ideal" criterion EPE.
$\l\approx5\times10^{-4}$ and $\l\approx7\times10^{-4}$ are selected by LoRP and GCV, respectively.
One again, like the previous example, LoRP selects a better $\l$ than GCV does.

%%%%%%%%%%%%%%%%%%%%%%%%%%%%%%%%%%%%%%%%%%%%%%%%%%%%%%%%%%%%%%%
\section{Comparison to Gaussian Bayesian Linear Regression}\label{secBayes}
%%%%%%%%%%%%%%%%%%%%%%%%%%%%%%%%%%%%%%%%%%%%%%%%%%%%%%%%%%%%%%%

We now consider LBFR from a
Bayesian perspective with Gaussian noise and prior, and compare it
to LoRP. In addition to the noise model as in PML, one also has to
specify a prior. Bayesian model selection (BMS) proceeds by
selecting the model that has largest evidence. In the special case
of LBFR with Gaussian noise and prior and a type II maximum likelihood estimate for the
noise variance, the expression for the evidence has a similar
structure as the expression of the loss rank.

%-------------------------------%
\paradot{Gaussian Bayesian LBFR / MAP}
%-------------------------------%
Recall from Sec.\ref{secLM} Ex.\ref{exLBFR} that $\F_d$ is the class
of functions $f_{\v w}(x)=\v w^\trp\v\phi(x)$ ($\v w\in\SetR^d$) that
are linear in feature vector $\v\phi$. Let
\beq\label{eqGauss}
  \Gauss_N(\v z|\v\mu,\s) \;:=\;
  {\exp(-\fr12(\v z-\v\mu)^\trp\s^{-1}(\v z-\v\mu))
   \over(2\pi)^{N/2}\sqrt{\det\s}}
\eeq
denote a general $N$-dimensional Gaussian distribution with mean
$\v\mu$ and covariance matrix $\s$.
We assume that observations $y$ are perturbed from $f_{\v w}(x)$ by
independent additive Gaussian noise with variance $\b^{-1}$ and zero
mean, i.e., the likelihood of $\v y$ under model $\v w$ is
$\P(\v y|\v w)=\Gauss_n(\v y|\Phi\v w,\b^{-1}I)$,
where $\Phi_{ia}=\v\phi_a(x_i)$.
A Bayesian assumes a prior (before seeing $\v y$) distribution on
$\v w$. We assume a centered Gaussian with covariance matrix $(\a
C)^{-1}$, i.e., $\P(\v w)=\Gauss_d(\v w|\v 0,\a^{-1}C^{-1})$.
From the prior and the likelihood one can compute the evidence and the posterior
\bqa\label{eqEv}
  \mbox{Evidence:}\qquad\quad\; \P(\v y) &=& \int\P(\v y|\v w)\P(\v w)d\v w
                                      \;=\; \Gauss_n(\v y|\v 0,\b^{-1}S^{-1})
\\ \nonumber
  \mbox{Posterior:}\qquad \P(\v w|\v y) &=& \P(\v y|\v w)\P(\v w)/P(\v y)
                                       \;=\; \Gauss_d(\v w|\v{\h w},A^{-1})
\eqa
\beq\label{eqBAMS}
  B:=\Phi^\trp\Phi, \quad
  A:=\a C+\b B, \quad
  M:=\b\Phi A^{-1}\Phi^\trp, \quad
  S:=I-M, \quad
\eeq
\beqn
  \v{\h w}:=\b A^{-1}\Phi^\trp\v y, \quad
  \v{\h y}:=\Phi\v{\h w}=M\v y
\eeqn
A standard Bayesian point estimate for $\v w$ for fixed $d$ is the
one that maximizes the posterior (MAP) (which in the Gaussian case
coincides with the mean)
$\v{\h w}=\arg\max_{\v w}\P(\v w|\v y)=\b A^{-1}\Phi^\trp\v y$.
For $\a\to 0$, MAP reduces to Maximum Likelihood (ML), which
in the Gaussian case coincides with the least squares regression of
Ex.\ref{exLBFR}. For $\a>0$, the regression matrix $M$ is not a
projection anymore.

%-------------------------------%
\paradot{Bayesian model selection}
%-------------------------------%
Consider now a family of models $\{\F_1,\F_2,...\}$. Here the $\F_d$
are the linear regressors with $d$ basis functions, but in general
they could be completely different model classes. All quantities in
the previous paragraph implicitly depend on the choice of $\F$,
which we now explicate with an index. In particular, the evidence
for model class $\F$ is $\P_\F(\v y)$.
BMS chooses the model class (here $d$) $\F$ of
highest evidence:
\beqn
  \F^{\text{BMS}} \;=\;\arg\max_\F\P_\F(\v y)
\eeqn
Once the model class $\F^{\text{BMS}}$ is determined, the MAP (or other)
regression function $f_{{\v w}_{\F^{\text{BMS}}}}$ or $M_{\F^{\text{BMS}}}$ are
chosen. The data variance $\b^{-1}$ may be known or estimated
from the data, $C$ is often chosen $I$, and $\a$ has to be chosen
somehow. Note that while $\a\to 0$ leads to a reasonable
MAP=ML regressor for fixed $d$, this limit cannot be used for BMS.

%-------------------------------%
\paradot{Comparison to LoRP}
%-------------------------------%
Inserting \eqref{eqGauss} into \eqref{eqEv} and taking the
logarithm we see that BMS minimizes
\beq\label{eqLLG1}
  -\log\P_\F(\v y) \;=\; \fr\b2\v y^\trp S\v y - \fr12\log\det S - \fr n2\log\fr\b{2\pi}
\eeq
w.r.t.\ $\F$. Let us estimate $\b$ by ML: We assume a broad prior
$\a\ll\b$ so that $\b{\partial S\over\partial\b}=O({\a\over\b})$ can
be neglected. Then $-{\partial\log\P_\F(\v y)\over\partial\b} =
\fr12\v y^\trp S\v y-{n\over 2\b}+O({\a\over\b}n)=0$ $\Leftrightarrow$
$\b\approx \h\b:=n/(\v y^\trp S\v y)$. Inserting $\h\b$
into \eqref{eqLLG1} we get
\beq\label{eqLLG2}
  -\log\P_\F(\v y) \;=\;
  \fr n2\log\v y^\trp S\v y - \fr12\log\det S - \fr n2\log\fr{n}{2\pi\e}
\eeq
Taking an improper prior $\P(\b)\propto\b^{-1}$ and integrating out
$\b$ leads for small $\a$ to a similar result. The last term in
\eqref{eqLLG2} is a constant independent of $\F$ and can be ignored.
The first two terms have the same structure as in linear LoRP
\eqref{eqLRL}, but the matrix $S$ is different.
In both cases, $\a$ act as regularizers, so we may minimize over
$\a$ in BMS like in LoRP. For $\a=0$ (which neither makes sense in
BMS nor in LoRP), $M$ in BMS coincides with $M$ of Ex.\ref{exLBFR},
but still the $S_0$ in LoRP is the square of the $S$ in BMS. For $\a>0$, $M$ of
BMS may be regarded as a regularized regressor as suggested in
Sec.\ref{secLoRP} (a), rather than a regularized loss function (b) used
in LoRP. Note also that BMS is limited to (semi)parametric regression,
i.e., does not cover the non-parametric kNN Ex.\ref{exKNN} and kernel
Ex.\ref{exKern}, unlike LoRP.

Since $B$ only depends on $\v x$ (and not on $\v y$), and all $\P$
are implicitly conditioned on $\v x$, one could choose $C=B$. In
this case, $M=\g\Phi B^{-1}\Phi^\trp$, with $\g={\b\over\a+\b}<1$
for $\a>0$, is a simple multiplicative regularization of projection
$\Phi B^{-1}\Phi^\trp$, and \eqref{eqLLG2} coincides with
\eqref{eqLRPa} for suitable $\a$, apart from an irrelevant additive
constant, hence minimizing \eqref{eqLLG2} over $\a$
also leads to \eqref{eqLRPamin}.

%%%%%%%%%%%%%%%%%%%%%%%%%%%%%%%%%%%%%%%%%%%%%%%%%%%%%%%%%%%%%%%
\section{Comparison to other Model Selection Schemes}\label{secOMS}
%%%%%%%%%%%%%%%%%%%%%%%%%%%%%%%%%%%%%%%%%%%%%%%%%%%%%%%%%%%%%%%

In this section we give a brief introduction to PML for (semi)parametric regression,
and its major instantiations,
AIC, BIC, and MDL principle,
whose penalty terms are all proportional to the number of parameters
$d$. The {\em effective number of parameters} is often much smaller than
$d$, e.g.,\ if there are soft constraints like in ridge regression. We
compare MacKay's trace formula \citep{MacKay:92} for Gaussian
Bayesian LBFR and Hastie's et al. trace formula \citep{Hastie:01}
for general linear regression with LoRP.

%-------------------------------%
\paradot{Penalized ML (AIC, BIC, MDL)}
%-------------------------------%
Consider a $d$-dimensional stochastic model class like the Gaussian
Bayesian linear regression example of Section \ref{secBayes}. Let
$\P_d(\v y|\v w)$ be the data likelihood under $d$-dimensional model
$\v w\in\SetR^d$. The maximum likelihood (ML) estimator for fixed
$d$ is
\beq\label{eqPMLwh}
  \v{\h w} \;=\; \arg\max_{\v w}\P_d(\v y|\v w)
  \;=\; \arg\min_{\v w}\{-\log \P_d(\v y|\v w)\}
\eeq
Since $-\log\P_d(\v y|\v w)$ decreases with $d$, we
cannot find the model dimension by simply minimizing over $d$
(overfitting). Penalized ML adds a complexity term to get
reasonable results
\beq\label{eqPMLdh}
  \h d \;=\; \arg\min_d\{-\log \P_d(\v y|\v{\h w}) + \mbox{Penalty}(d) \}
\eeq
The penalty introduces a tradeoff between the first and second
term with a minimum at $\h d<\infty$. Various penalties have been
suggested: AIC \citep{Akaike:73}
uses $d$, BIC \citep{Schwarz:78}
and the (crude) MDL \citep{Rissanen:78,Gruenwald:04} use $\fr
d2\log n$ for Penalty$(d)$.
There are at least {\em three important conceptual differences} to LoRP:
\begin{itemize}\parskip=0ex\parsep=0ex\itemsep=0ex
\item In order to apply PML one needs to specify not only a class
of regression functions, but a full probabilistic model $\P_d(\v y|\v w)$,
\item PML ignores or at least does not tell how to incorporate
a potentially given loss-function,
\item PML is mostly limited to selecting between (semi)parametric models.
\end{itemize}

We discuss two approaches to the last item in the remainder of this
section (where AIC, BIC, and MDL are not directly applicable): %
(a) for non-parametric models like kNN or kernel regression, or %
(b) if $d$ does not reflect the ``true'' complexity of the model.
\iffalse
For instance, ridge regression can work even for $d$ larger than
$n$, because a penalty pulls most parameters towards (but not
exactly to) zero. %
\fi
\cite{MacKay:92} suggests an expression for the effective
number of parameters $d_{e\!f\!f}$ as a substitute for $d$ in case
of (b), while \cite{Hastie:01} introduce another expression which is applicable for
both (a) and (b).

%-------------------------------%
\paradot{The trace penalty for parametric Gaussian LBFR}
%-------------------------------%
We continue with the Gaussian Bayesian linear regression example
(see Section \ref{secBayes} for details and notation). Performing
the integration in \eqref{eqEv}, \citet[Eq.(21)]{MacKay:92}
derives the following expression for the Bayesian evidence for $C=I$
\bqa\label{eqLLMK}
  -\log\P(\v y) &=&
  (\a \h E_W+\b \h E_D) + (\fr12\log\det A -\fr d2\log\a) - \fr n2\log\fr\b{2\pi}
\\ \nonumber
  \h E_D &=& \fr12\|\Phi\v{\h w}-\v y\|_2^2, \quad
  \h E_W =\fr12\|\v{\h w}\|_2^2
\eqa
(the first bracket in \eqref{eqLLMK} equals $\fr\b2\v y^\trp S\v y$ and
the second equals $-\fr12\log\det S$, cf.\ \eqref{eqLLG1}).
Minimizing \eqref{eqLLMK} w.r.t.\ $\a$ leads
to the following relation:
\beqn
  0 \;=\; \textstyle{-\partial\log\P(\v y)\over\partial\a} \;=\;
  \h E_W +\fr12\tr A^{-1}-\fr d{2\a} \qquad
 ({\partial\over\partial\a}\log\det A=\tr A^{-1})
\eeqn
He argues that
$\a\|\v{\h w}\|_2^2$ corresponds to the effective number of
parameters, hence
\beq\label{eqdeffMcK}
  d^{\text{McK}}_{e\!f\!f} \;:=\; \a\|\v{\h w}\|_2^2
  \;=\; 2\a \h E_W \;=\; d-\a\tr A^{-1}
\eeq

%-------------------------------%
\paradot{The trace penalty for general linear models}
%-------------------------------%
We now return to general linear regression $\v{\h y}=M(\v x)\v y$
\eqref{eqM}. LBFR is a special case of a projection matrix $M=M^2$
with rank $d=\tr M$ being the number of basis functions. $M$ leaves
$d$ directions untouched and projects all other $n-d$ directions to
zero. For general $M$, \citet[Sec.5.4.1]{Hastie:01}
argue to regard a direction that is only somewhat shrunken, say by a
factor of $0<\b<1$, as a fractional parameter ($\b$ degrees of
freedom). If $\b_1,...,\b_n$ are the shrinkages = eigenvalues
of $M$, the effective number of parameters could be defined as
\cite[Sec.7.6]{Hastie:01}
\beqn
  d^{\text{HTF}}_{e\!f\!f} \;:=\; \sum_{i=1}^n \b_i \;=\; \tr M,
\eeqn
where HTF stands for Hastie-Tibshirani-Friedman,
which generalizes the relation $d=\tr M$ beyond projections.
For MacKay's $M$ \eqref{eqBAMS}, $\tr M=d-\a\tr A^{-1}$,
i.e., $d^{\text{HTF}}_{e\!f\!f}$ is consistent with and generalizes
$d^{\text{McK}}_{e\!f\!f}$.

%-------------------------------%
\paradot{Problems}
%-------------------------------%
Though nicely motivated, the trace formula is not without problems.
First, since for projections, $M=M^2$, one could have
argued equally well for $d^{\text{HTF}}_{e\!f\!f}=\tr M^2$. Second, for kNN we have $\tr
M=\fr nk$ (since $M$ is $\fr 1k$ on the diagonal), which does not
look unreasonable. Consider now kNN',
which is defined as follows: we average over the $k$
nearest neighbors {\em excluding} the closest neighbor. For
sufficiently smooth functions, kNN' for suitable $k$ is still a
reasonable regressor, but $\tr M=0$ (since $M$ is zero on the
diagonal). So $d^{\text{HTF}}_{e\!f\!f}=0$ for kNN', which makes no sense
and would lead one to always select the $k=1$ model.

%-------------------------------%
\paradot{Relation to LoRP}
%-------------------------------%
In the case of kNN', $\tr M^2$ would be a better estimate for the
effective dimension. In linear LoRP, $-\log\det S_\a$ serves as
complexity penalty. Ignoring the nullspace of
$S_0=(I-M)^\trp(I-M)$ \eqref{defSa}, we can Taylor expand
$-\fr12\log\det S_0$ in $M$
\beqn
  -\fr12\log\det S_0 \;=\; -\tr\log(I\!-\!M)
  \;=\; \sum_{s=1}^\infty \fr1s\tr(M^s)
  \;=\; \tr M + \fr12\tr M^2 + ...
\eeqn
For BMS \eqref{eqLLG2} with $S=I-M$ \eqref{eqBAMS} we get half of
this value. So the trace penalty may be regarded as a leading order
approximation to LoRP. The higher order terms prevent
peculiarities like in kNN'.

%-------------------------------%
\paradot{Coding/MDL interpretation of LoRP}
%-------------------------------%
The basic idea of MDL is as follows \citep{Gruenwald:04}:
``The goal of statistical inferences may be cast as trying to find
regularity in the data. `Regularity' may be identified with `ability
to compress'. MDL combines these two insights by {\em viewing
learning as data compression}: it tells us that, for a given set of
hypotheses $\cal H$ and data set $D$, we should try to find the
hypothesis or combination of hypotheses in $\cal H$ that compress
$D$ most.''

The standard incarnation of (crude) MDL is as follows: If $H$ is a
stochastic model of (discrete) data $D$, we can code $D$ (by
Shannon-Fano) in $\lceil-\lb\P(D|H)\rceil$ bits. If we have a class
of models $\cal H$, we also have to code $H$ (somehow in, say, $L(H)$
bits) in order to be able to decode $D$. MDL chooses the hypothesis
$H^{\MDL}=\arg\min_{H\in\cal H}\{-\lb\P(D|H)+L(H)\}$ of minimal
two-part code. For instance, if $\cal H$ is the class of all
polynomials of all degrees with each coefficient coded
to $\fr12\lb n$ bits (i.e., $O(n^{-1/2})$ accuracy)
and we condition on $x$, i.e., $D\leadsto\v y|\v x$, MDL
takes the form \eqref{eqPMLwh} and \eqref{eqPMLdh}, i.e., $H^{\MDL}=(\v{\h w},\h d)$.

We now give LoRP (for discrete $D$) a data compression/MDL interpretation.
For simplicity, we will first assume that all loss values are
different, i.e., if $\L_r(\v y'|\v x)\neq\L_r(\v y''|\v x)$ for $\v
y'\neq\v y''$ (adding infinitesimal random noise to $\L_r$ easily
ensures this). In this case, $\Rank_r(\cdot|\v x):\Y^n\to\SetN$ is an order
preserving bijection, i.e., $\Rank_r(\v y'|\v x)<\Rank_r(\v y''|\v
x)$ iff $\L_r(\v y'|\v x)<\L_r(\v y''|\v x)$ with no gaps in the
range of $\Rank_r(\cdot|\v x)$.

Phrased differently, $\Rank_r(\cdot|\v x)$ codes each $\v y'\in\Y^n$
as a natural number $m$ in increasing loss-order. The natural number
$m$ can itself be coded in $\lceil\lb m\rceil$ bits (using plain not prefix coding). Let us call
this code of $\v y'$ the {\em Loss Rank Code} (LRC). LRC has a nice
characterization: LRC is the shortest loss-order preserving code.
Ignoring the rounding, the {\em Length} of LRC$_r(\v y'|\v x)$ is
$\LR_r(\v y'|\v x)$:

%-------------------------------%
\begin{proposition}[Minimality property]\label{thmMP}
%-------------------------------%
If all loss values are different, i.e., if
\beqn
  \L_r(\v y'|\v x)\neq\L_r(\v y''|\v x)
  \mbox{ for all }\v y'\neq\v y''
\eeqn
then the loss rank (code) of $\v y$ is the smallest/shortest among all
loss-order preserving rankings/codes $C$ in the sense that
\bqan
  \Rank(\v y) &=& \min\{C(\v y) \,:\, C\in\Y^n\!\to\!\SetN \,\wedge\, (\star)\,\}
\\
  \lfloor\LR(\v y)/\log 2\rfloor &=& \min\{\text{\rm Length}(C(\v y))\,:\,
    C\in\Y^n\!\to\!\{0,1\}^* \,\wedge\, (\star)\,\}
\\
  (\star) &:=& [ \L(\v y')<\L(\v y'') \Leftrightarrow C(\v y')<C(\v y''),\,\forall\v y',\v y'']
\eqan
\end{proposition}

The proof follows from the fact that if a discrete injection (code)
is order preserving, there exists a ``smallest'' one without gaps in
the range. So LoRP minimizes the Loss Rank Code, where LRC itself is
the shortest among all loss-order preserving codes.
From this perspective, LoRP is just a different (non-stochastic,
non-parametric, loss-based) incarnation of MDL.
The MDL philosophy provides a justification of LoRP \eqref{eqLRD}, its
regularization \eqref{eqLa}, and loss function selection (Section
\ref{secLFS}). This identification should also allow to apply or
adapt the various consistency results of MDL, implying that LoRP is
consistent under some mild conditions.

If some losses are equal, $\Rank_r(\cdot|\v x):\Y^n\to\SetN$ still
preserves the order $\leq$, but the mapping is neither surjective
nor injective anymore.

%-------------------------------%
\paradot{Large regression classes $\R$}
%-------------------------------%
The classes $\R$ of regressors we considered so far were discrete
and ``small'', often indexed by an integer complexity index (like
$k$ in kNN or $d$ in LBFR). But large classes are also thinkable.

As an extreme case, consider the class of {\em all} regressors.
Clearly, there is an $r=r_D$ which ``knows'' $D$ and perfectly fits
$D$ ($r(x_i|D)=y_i,\ \forall i$), but is the worst possible on all
other $D'$ ($r(x_i|D')=\infty,\ \forall i,\ \forall D'\neq D$).
This $r$ has (discrete) Rank 1, so is best according to LoRP.
So if $\R$ is too large, LoRP can overfit too.

Consider a more realistic example by not taking {\em all} of the
first $d$ basis functions in LBFR, but selecting {\em some} basis
functions $\phi_{i_1},...,\phi_{i_d}$, i.e., $\R$ is indexed by $d$
integers, and $d$ may be variable too.

One solution approach is to group more regressors in $\R$ into one
function class $\F$, e.g.,\ the class of functions
$\F_{k,d}=\{w_1\phi_{i_1}+...w_d\phi_{i_d}:\v w\in\SetR^d,\, 1\leq
i_1<...<i_d\leq k\}$ that are linear in $d$ of the first $k$ bases.
Now $\R$ is a small class indexed by $d$ and $k$ only.

Looking at the coding interpretation of $\LR_r$ and the MDL
philosophy, suggests to assign a code to $r\in\SetR$ in order to get
a complete code for $D$:
\beqn
  r^{best} \;=\; \arg\min_r \{\LR_r(\v y|\v x) + L(r)\}
\eeqn
where $r$ is the length of a code for $r$ (given $\R$). For
$\R\simeq\SetN$ a single integer has to be coded, e.g.,\ $k$ in
$L(r)=L(k)\approx\lb k$ bits, which can usually be safely
dropped/ignored. For more complex classes like the (ungrouped) LBFR subset
selection above, $L(r)=L(i_1,...,i_d,d)\approx d\lb k+\lb d$ can
become important.

%%%%%%%%%%%%%%%%%%%%%%%%%%%%%%%%%%%%%%%%%%%%%%%%%%%%%%%%%%%%%%%
\section{Loss Functions and their Selection}\label{secLFS}
%%%%%%%%%%%%%%%%%%%%%%%%%%%%%%%%%%%%%%%%%%%%%%%%%%%%%%%%%%%%%%%

%-------------------------------%
\paradot{General additive loss}
%-------------------------------%
Linear LoRP $\v{\h y}=M(\v x)\v y$ of Section \ref{secLM} can easily
be generalized to non-quadratic loss. Let us consider the
$\rho>0$ loss
\bqan
  \L_M(\v y|\v x) &:=& \textstyle (\sum_{i=1}^n(y_i-\h y_i)^\rho)^\frs1\rho
  \;=\; \|\v y-\v{\h y}\|_\rho \;=\; \|(I\!-\!M)\v y\|_\rho
\\
  V(L) &=& \{\v y'\in\SetR^n:\|(I\!-\!M)\v y'\|_\rho\leq L\}
  \;=\; \{(I\!-\!M)^{-1}\v z\in\SetR^n:\|\v z\|_\rho\leq L\} %,\v z\in\SetR^n
\\
  \mbox{Let}\quad v_n^\rho &:=& |\{\v z\in\SetR^n:\|\v z\|_\rho\leq 1\}|
  \;=\; \textstyle 2^n\prod_{i=1}^{n-1}{i\over\rho}!{1\over\rho}!/{i+1\over\rho}!,
\eqan
where $\frac{i}{\rho}!:=\Gamma(\frac{i}{\rho}+1)$, be the volume of
the unit $d$-dimensional $\rho$-norm ``ball''. Since $V(L)$ is a
linear transformation of this ball with transformation matrix
$(I-M)^{-1}$ and scaling $L$, we have $|V(L)|=v_n^\rho
L^n/\det(I-M)$, hence
\beq\label{eqLRLrho}
  \LR_M(\v y|\v x) \;=\; \log|V(\L_M(\v y|\v x))|
  \;=\; n\log\|(I\!-\!M)\v y\|_\rho - \log\det(I\!-\!M) + \log v_n^\rho
\eeq
For the $\rho=2$ norm, \eqref{eqLRLrho} reduces to $\LR_M^0$
\eqref{eqLRL1}.
Note that $\L_M:=g(\|\v y-\v{\h y}\|_\rho)$ leads to the
same result \eqref{eqLRLrho} for any monotone increasing $g$, i.e., only the order of the loss matters, not its absolute
value.
More generally $\L_M=g(\sum_i h(y_i-\h y_i))$ for any $h$ implies
\bqan
  \LR_M(\v y|\v x) &=& \textstyle n\log v_n^h(\sum_i h(y_i-\h y_i)) - \log\det(I\!-\!M), \qmbox{where}
\\
  v_n^h(l) &:=& \textstyle |\{\v z\in\SetR^n:\sum_i h(z_i)\leq l\}|^{\frs1n}
\eqan
is a one-dimensional function of $l$ (independent $D$ and $M$), once
to be determined (e.g.,\ $v_n^h(l)=l\cdot (v_n^\rho)^{\frs1n}\propto l$ for $\rho$-norm loss).
Regularization may be performed by $M\leadsto\gamma M$
with optimization over $\gamma<1$.

%-------------------------------%
\paradot{Loss-function selection}
%-------------------------------%
In principle, the loss function should be part of the problem
specification, since it characterizes the ultimate goal.
For instance, whether a test should more likely
classify a healthy person as sick than a sick person as healthy,
depends on the severity of a misclassification (loss) in each
direction.
In reality, though, having to specify the loss function can be a
nuisance. Sure, the loss has to respect some general features,
e.g.,\ that it increases with the deviation of $\h y_i$ from
$y_i$. Otherwise it is chosen by convenience or rules of thumb,
rather than by elicitation of the real goal, for instance preferring the
Euclidean norm over $\rho\neq 2$ norms.
If we subscribe to the procedure of {\em choosing} the loss
function, we could ask whether this may be done in a more principled
way. Consider a (not too large) class of loss functions $\L^\a$,
indexed by some parameter $\a$. For instance, $\L^\a=\|\v y-\v{\h
y}\|_\a$ from the previous paragraph. The regularized loss
\eqref{eqLa} also constitutes a class of losses. In this case we
minimized over the regularization parameter $\a$. This suggests to
choose in general the loss function that has minimal loss rank
$\LR_r^\a$. The justifications are similar to the ones for
minimizing $\LR_r^\a$ w.r.t.\ $r$. Note that the term
$\log v_n^\rho$ cannot be dropped anymore, unlike in \eqref{eqLRL}.

%%%%%%%%%%%%%%%%%%%%%%%%%%%%%%%%%%%%%%%%%%%%%%%%%%%%%%%%%%%%%%%
\section{Self-Consistent Regression}\label{secSCR}
%%%%%%%%%%%%%%%%%%%%%%%%%%%%%%%%%%%%%%%%%%%%%%%%%%%%%%%%%%%%%%%

So far we have considered only ``on-data'' regression. LoRP only
depends on the regressor $r$ on data $D$ and not on
$x\not\in\{x_1,...,x_n\}$.
We now construct canonical regressors for off-data $x$ from
regressors given only on-data. First, this may ease the
specification of the regression functions, second, it is a canonical
way for interpolation (LoRP can't distinguish between $r$ that are
identical on $D$), and third, we show that many standard regressors
(kNN, Kernel, LBFR) are self-consistent in the sense that they are
canonical. We limit our exposition to linear regression.

%-------------------------------%
\paradot{Off-data regression}
%-------------------------------%
A linear regressor is completely determined by the $n$ functions
$m_j$ \eqref{eqmj}, but not by the matrix function $M$ \eqref{eqM}.
Indeed, two sets $\{m_j\}$ and $\{m'_j\}$ that coincide on $D=(\v
x,\v y)$, i.e.\ $m_j(x_i|\v x)=m'_j(x_i|\v x)\,\forall i,j$ but
possibly differ for $x\not\in\v x$, lead to the same matrix
$M_{ij}(\v x)=m_j(x_i|\v x)=m'_j(x_i|\v x)$. LoRP has the advantage
of only depending on $M$, but this also means that it cannot
distinguish between an $m_j$ that behaves well on $x\not\in\v x$ and
one that, e.g.,\ wildly oscillates outside $\v x$.

Typically, the $m_j$ are given and, provided the model complexity is
chosen appropriately e.g.\ by LoRP, they properly interpolate $\v
x$. Nevertheless, a canonical extension from $M$ to $m_j$ would
be nice. In this way LoRP would not be vulnerable to bad $m_j$, and
we could interpolate $D$ (predict $y$ for any $x\in\X$) even without
$m_j$ given a-priori.

We define a self-consistent regression scheme based only on $M$ (for all
$n$). We ask for an estimate $\h y$ of $y$ for $x\not\in\v x$. We
add a virtual data point $(x_0,y_0)$ to $D$, where $x_0=x$. If we
knew $y_0=y$ we could estimate $\h y_0=r(x_0|\{(x_0,y_0)\}\cup D)$,
but we don't know $y_0$. But we could require a self-consistency
condition, namely that $\h y_0=y_0$ for $x_0\not\in\v x$.

%-------------------------------%
\begin{definition}[canonical and self-consistent regressors]
%-------------------------------%
Let $M'_{ij}(\v x')_{0\leq i,j\leq n}$ be the regression matrix for the
data set $D'=\{(x_0,y_0)\}\cup D=((x_0,\v
x),(y_0,\v y))=(\v x',\v y')$ of size $n+1$.
\begin{itemize}\parskip=0ex\parsep=0ex\itemsep=0ex

\item[(i)] A linear regressor $\tilde y_0=\tilde r(x_0|D)$ is called a
canonical regressor for $M'$ if the consistency condition
$\tilde y_0=r(x_0|D')\equiv\sum_{j=0}^n M'_{0j}y_j$ holds $\forall x_0,D$.

\item[(ii)] A regressor $r$ is called
self-consistent if $\tilde r=r$, i.e.\ if \\
$r(x_0|\{(x_0,r(x_0|D))\}\cup D)=r(x_0|D)$ $\forall x_0,D$.

\item[(iii)] A class of regressors $\R=\{r\}$ is called self-consistent if
$\tilde R=\{\tilde r\}\subseteq\R$.
\end{itemize}
\end{definition}

We denote the solution of the self-consistency condition
$y_0=\sum_{j=0}^n M'_{0j}y_j$ by $\tilde
y_0$. So we have to solve
\beqn
  \tilde y_0 \;=\; \sum_{j=1}^n M'_{0j}y_j+M'_{00}\tilde y_0
  \quad\Longrightarrow\quad
  \tilde y_0 \;=\; {\sum_{j=1}^n M'_{0j}y_j\over 1-M'_{00}}
  \;=\; {\sum_{j=1}^n M'_{0j}y_j\over\sum_{j=1}^n M'_{0j}}
\eeqn
where the last equality only holds if $\sum_{j=0}^n M'_{0j}=1$,
which is often the case, in particular for kNN and Kernel regression,
but not necessarily for LBFR.

%-------------------------------%
\begin{proposition}[canonical regressor]
%-------------------------------%
The linear regressor
\beqn
  y_0 \;=\; \tilde r(x_0|D) \;:=\; \sum_{j=1}^n \tilde
  m_j(x_0|\v x)y_j, \qmbox{where}
  \tilde m_j(x_0|\v x):={M'_{0j}(\v x')\over 1-M'_{00}(\v x')}
\eeqn
is the unique canonical regressor for $M'$ (if $M'_{00}<1$).
\end{proposition}

%-------------------------------%
\fexample{exSCKNN}{self-consistent kNN, $\uparrow$Ex.\ref{exKNN}}{
%-------------------------------%
$M'_{0j}(\v x')=\fr1k$ for $x_j\in\N'_k(x_0)$ and 0 else.
The $k$ nearest neighbors $\N'_k(x_0)$ of $x_0$ among $\v x'$
consist of $x_0$ and the $k-1$
nearest neighbors $\N_{k-1}(x_0)=:J$ of $x_0$ among $\v x$, i.e.\
$\N'_k(x_0)=\{x_0\}\cup\N_{k-1}(x_0)$. Hence
\beqn
  \tilde y_0 \;=\; {\sum_{j=1}^n M'_{0j}y_j\over\sum_{j=1}^n M'_{0j}}
  \;=\; {\sum_{j\in J} \fr1k y_j\over\sum_{j\in J} \fr1k}
  \;=\; \sum_{j\in J} \fr1{k-1} y_j
  \;=\; \sum_{j=1}^n M_{0j}^{(k-1)}y_j
  \;=\; r_{k-1}(x_0|D)=\h y_0
\eeqn
Canonical kNN is equivalent to standard (k--1)NN, so the class of
canonical kNN regressors coincides with the standard kNN class.
} % End of Example

%-------------------------------%
\fexample{exSCKern}{self-consistent kernel}{
%-------------------------------%
\beqn
  M'_{0j}(\v x') \;=\; {K(x_0,x_j)\over\sum_{j=0}^n K(x_0,x_j)}
  \quad\Longrightarrow\quad
  \tilde y_0
%  \;=\; {\sum_{j=1}^n M'_{0j}y_j\over\sum_{j=1}^n M'_{0j}}
  \;=\; {\sum_{j=1}^n K(x_0,x_j)y_j\over\sum_{j=1}^n K(x_0,x_j)}
  \;=\; r(x_0|D) \;=\; \h y_0
\eeqn
Canonical kernel regression coincides with the standard
kernel smoother.
} % End of Example

%-------------------------------%
\fexample{exSCLBFR}{self-consistent LBFR}{
%-------------------------------%
\bqan
  B' &=& \sum_{i=0}^n\v\phi(x_i)\v\phi(x_i)^\trp = B+\v\phi(x_0)\v\phi(x_0)^\trp
\\
  \Rightarrow\quad
  M'_{0j} &=& \v\phi(x_0)^\trp B'\!\,^{-1}\v\phi(x_j)
  \;=\; \v\phi(x_0)^\trp\bigg[B^{-1}-{B^{-1}\v\phi(x_0)\v\phi(x_0)^\trp B^{-1}\over
    1+\v\phi(x_0)^\trp B^{-1}\v\phi(x_0)}\bigg]\v\phi(x_j)
\\
  &=& M_{0j}-{M_{00}M_{0j}\over 1+M_{00}}
  \;=\; {M_{0j}\over 1+M_{00}}
  \quad\Rightarrow\quad 1-M'_{00} = {1\over 1+M_{00}}
\eqan
In the first line we used the Sherman-Morrison formula for inverting $B'$.
In the second line we defined $M_{0j}=\v\phi(x_0)^\trp B^{-1}\v\phi(x_j)$,
extending $M$.
\beqn
  \Rightarrow\quad \tilde y_0 \;=\; {\sum_{j=1}^n M'_{0j}y_j\over 1-M'_{00}}
  \;=\; \sum_{j=1}^n M_{0j}y_j \;=\; \sum_{j=1}^n m_j(x_0,\v x)y_j \;=\; \h y_0
\eeqn
Canonical LBFR coincides with standard LBFR.
} % End of Example

%-------------------------------%
\begin{proposition}[self-consistent regressors]
%-------------------------------%
Kernel regression and linear basis function regression are
self-consistent. kNN is not self-consistent but the class
of kNN regressors $\R=\{r_{\text{kNN}}:k\in\SetN\}$ is self-consistent.
\end{proposition}

To summarize, we expect LoRP to select good regressors with proper
interpolation behavior for canonical and self-consistent regressors.

%%%%%%%%%%%%%%%%%%%%%%%%%%%%%%%%%%%%%%%%%%%%%%%%%%%%%%%%%%%%%%%
\section{Nearest Neighbors Classification}\label{secKNN}
%%%%%%%%%%%%%%%%%%%%%%%%%%%%%%%%%%%%%%%%%%%%%%%%%%%%%%%%%%%%%%%

We now consider k-nearest neighbors classification in more detail.
In order to get more insight into LoRP we seek a case that allows
analytic solution. In general, the determinant $\det S_\a$ cannot be
computed analytically, but for $\v x$ lying on a hypercube of the
regular grid $\X=\SetZ^d$ we can. We derive exact expressions, and
consider the limits $n\to\infty$, $k\to\infty$, and $d\to\infty$.

%-------------------------------%
\paradot{kNN on one-dimensional grid}
%-------------------------------%
We consider the $d=1$ dimensional case first.
We assume $\v x=(1,2,3,...,n)$, a circular metric
$d(x_i,x_j)=d(i,j)=\min\{|i-j|,n-|i-j|\}$, and odd $k\leq n$.
The kNN regression matrix
\iffalse
$M_{ij}=b_{i-j}$ with $b_{i-j}=1$ if $d(i,j)\leq {k-1\over 2}$ and 0 else.
\else
\beqn
  M_{ij}=b_{i-j} \qmbox{with} b_{i-j}=\fr1k \qmbox{if}
  d(i,j)\leq \fr{k-1}{2} \qmbox{and} 0 \qmbox{otherwise}
\eeqn
\fi
is a diagonal-constant
(Toeplitz) matrix with circularity property $b_{i-j}=b_{i-j+n}$.
%called circulant matrix.
For instance, for $k=3$ and $n=5$
\beqn
  M = {1\over 3}
  \left(\scriptsize\arraycolsep=2pt
    \begin{array}{ccccc}
      1 & 1 & 0 & 0 & 1 \\
      1 & 1 & 1 & 0 & 0 \\
      0 & 1 & 1 & 1 & 0 \\
      0 & 0 & 1 & 1 & 1 \\
      1 & 0 & 0 & 1 & 1 \\
    \end{array}
  \right)
\eeqn
For every circulant matrix, the eigenvectors $\v v^1,...,\v v^n$ are
waves $v_j^l=\t^{jl}$ with $\t=\e^{2\pi\sqrt{-1}/n}$. The
eigenvalues are the fourier transform $\h b_l=\sum_{j=1}^n
b_j\t^{-jl}$ of $\v b$, since $\sum_j M_{ij}v_j^l =
\sum_j b_{i-j}\t^{jl} = \sum_j b_j\t^{(i-j)l} = v_i^l\sum_j
b_j\t^{-jl} = \h b_l v_i^l$, where we exploited circularity of $\v b$
and $\t^{jl}$. For $M_{\text{kNN}}$ in particular we get
\beqn
  \h b_l
  \;\mathop=_{\displaystyle\mathop{\rule{0ex}{3ex}
    \uparrow}_{\rule{0ex}{2ex}\makebox[0ex]{\footnotesize
    circularity}}}\;
  {1\over k}\sum_{\nq j=-{k-1\over 2}\nq}^{k-1\over 2}\t^{-jl}
  \;\mathop=_{\displaystyle\mathop{\rule{0ex}{3ex}
    \uparrow}_{\rule{0ex}{2ex}\makebox[0ex]{\footnotesize
    $\qquad$ geometric sum}}}\;
  {1\over k}{\t^{lk/2}-\t^{-lk/2}\over\t^{l/2}-\t^{-l/2}}
  \;\mathop=_{\displaystyle\mathop{\rule{0ex}{3ex}
    \uparrow}_{\rule{0ex}{2ex}\makebox[0ex]{\footnotesize
    insert $\t$}}}\;
  {\sin(\pi lk/n)\over k\sin(\pi l/n)} \;<\; 1
  \qmbox{for} l\neq n
\eeqn
and $\h b_n=1$. The only 1-vector $\v v^n=\bf 1$ corresponds to a
constant shift $y_i\leadsto y_i+c$ under which kNN (like many other
regressors) is invariant. Instead of regularizing LoRP with $\a>0$
we can restrict $V(L)\subset \SetR^n$ to the space orthogonal to $\v
v^n$, which means dropping $\h b_n=1$ in the determinant.
Intuitively, since this invariant direction is the same for all $k$,
we can drop the same additive infinite constant from $\LR$ for every
$k$, which is irrelevant for comparisons (formally we should
compute $\lim_{\a\to 0}\{\LR_{k_1}^\a-LR_{k_2}^\a\}$). The exact
expression for the restricted log-determinant (denoted by a prime)
is
\beqn
  -\fr12\log\det\nolimits'\!\!S_0
  \;=\; -\log\det\nolimits'\!(\I\!-\!M)
  \;=\; -\sum_{l=1}^{n-1}\log(1\!-\!\h b_l)
  \;=:\;\fr n k c_{nk}^1 \;=\; c_{nk}^1\tr M
\eeqn
For large $n$ (and large $k$) the expression can be simplified. The
exact, large $n$, and large $k\ll n$ expressions are
\bqan
  c_{nk}^1 &=& -{k\over n}\sum_{l=1}^{n-1}\log\Big(1-{\sin(\pi lk/n)\over k\sin(\pi l/n)}\Big)
\\
  c_{\infty k}^1 &=& -{k\over\pi}\int_{-\pi/2}^{\pi/2}\log\Big(1-{\sin(kz)\over k\sin(z)}\Big)dz
  \qquad\qquad \left({z=\pi l/n\text{ for } l<\fr n2\atop z=\pi l/n-\pi\text{ else}}\right)
\\
  c_{\infty\infty}^1 &=& -{1\over \pi}\int_{-\infty}^\infty\log\Big(1-{\sin t\over t}\Big)\,dt
  \;\dot=\; 3.202
  \qquad (t=kz, \sin(z)\sim z)
\eqan
Further, $c_{\infty 3}^1 = 3\log 3\dot = 3.295$. %3.295836867
Since $c_{\infty k}^1$ is decreasing in $k$, $c_{\infty k}^1$ equals
$3.2$ within $3\%$ for all $k$.

%-------------------------------%
\paradot{kNN on $d$-dimensional grid}
%-------------------------------%
We now consider $\v x=\X^d=\{1,...,n_1\}^d$ on a $d$-dimensional
complete hypercube grid with $n=n_1^d$ points and Manhattan distance
$d(x_{\v i},x_{\v j})=d(\v i,\v j)=\sum_{a=1}^d d_1(i_a,j_a)$ for all
$x_{\v i}=\v i\in\X^d$ and $x_{\v j}=\v j\in\X^d$, where
$d_1$ is the one-dimensional circular distance defined above (so
actually $\X^d$ is a discrete torus). For $k=k_1^d$, the
neighborhood $\N_k(x)$ of $x$ is a cube of side-length $k_1$. In
this case, $M=M_1\otimes...\otimes M_1$ is a $d$-fold tensor product
of the 1d k$_1$NN matrices $M_1$ of sample size $n_1$. The
eigenvectors of $M$ are $\v v^{l_1}\otimes...\otimes\v v^{l_d}$ with
eigenvalues $\h b_{l_1}\cdot...\cdot\h b_{l_d}$.
We get
\bqa\label{eqTaylorM}
  -\log\det\nolimits'(\I-M)
    &=& -\sum_{l_1=1}^{n_1-1}...\sum_{l_d=1}^{n_d-1}
         \log(1-\h b_{l_1}\!\cdot...\cdot\h b_{l_d})
\\ \nonumber
  &\;\stackrel{n\gg k\to\infty}\longrightarrow\;&
  -{1\over\pi^d}\int_{\SetR^d}\log\Big(1-\prod_{a=1}^d{\sin t_a\over t_a}\Big)d^d\v t
  \;=:\; \fr n k c_{\infty\infty}^d
\eqa
For instance, for $d=2$, numerical integration gives
$c_{\infty\infty}^2\dot=2.2$ % verify !
compared to $3.2$ in one dimension. For higher dimensions,
evaluation of the $d$-dimensional integral becomes cumbersome, and
we resort to a different approximation.

%-------------------------------%
\paradot{Taylor series in $M$}
%-------------------------------%
We can also (not only for kNN) expand $\log\det S_0$ in a Taylor series
in $M$:
\bqan
  -\log\det\nolimits'(\I\!-\!M) &=& -\tr'\log(\I\!-\!M)
  \;=\; \sum_{s=1}^\infty\fr1s\tr'(M^s)
\\
  &=& \sum_{s=1}^\infty\fr1s(\tr'M_1^s)^d
  \;=\; \fr nk\sum_{s=1}^\infty\fr1s(A_{n_1 k_1 s})^d
  \;=:\; \fr nk c_{nk}^d
\eqan
where we used $\tr(A\otimes B)=\tr(A)\cdot\tr(B)$ and $(A\otimes
B)^s=A^s\otimes B^s$ and defined
\beqn
  A_{n_1 k_1 s} \;:=\; {k_1\over n_1}\tr'(M_1^s)
  \;=\; {k_1\over n_1}\sum_{l=1}^{n_1-1} (\h b_l)^s
  \;\stackrel{n\gg k\to\infty}\longrightarrow\;
  {1\over\pi}\int_{-\infty}^\infty\Big({\sin t\over t}\Big)^s dt
\eeqn
The one-dimensional integral can be expressed as a finite sum with
$s$ terms or evaluated numerically. For any $n$ and $k$ one
can show that $A_{nk1}=A_{nk2}=1>A_{nks}$ for $s>2$. So the expansion
above is useful for large $d$. Note also that $c_{nk}^d$ is monotone
decreasing in $d$. For $d\to\infty$ we have
\beqn
  c_{nk}^\infty \;=\; \sum_{s=1}^\infty \fr1s(A_{nks})^\infty
  \;=\; 1+\fr12+0+... \;=\; \fr32
\eeqn
i.e.\ $c_{nk}^d$ decreases monotone in $d$ from about 3.2 to $\fr32$.

The practical implication of this observation, though, is limited,
since $k=k_1^d\to\infty$ is actually not fixed for $d\to\infty$.
Indeed, in practical high-dimensional problems, $k\ll  n\ll 3^d$,
but in our grid example $k=k_1^d\geq 3^d$. Real data do not form
full grids but sparse neighborhoods if $d$ is large.

%%%%%%%%%%%%%%%%%%%%%%%%%%%%%%%%%%%%%%%%%%%%%%%%%%%%%%%%%%%%%%%
\section{Conclusion and Outlook}\label{secMisc}
%%%%%%%%%%%%%%%%%%%%%%%%%%%%%%%%%%%%%%%%%%%%%%%%%%%%%%%%%%%%%%%
%[tran]
We introduced a new method, the Loss Rank Principle, for model
selection. The loss rank of a model is defined as the number of
other data that fit the model better than the training data. The
model chosen by LoRP is the one of smallest loss rank. The loss rank
has an explicit expression in case of linear models.
Model consistency and asymptotic efficiency of LoRP were considered. The
numerical experiments suggest that LoRP works well in practice.
A comparison between LoRP and other methods for model selection was
also presented.

In this paper, we have only scratched at the surface of LoRP.
LoRP seems to be a promising principle with a lot of
potential, leading to a rich field. In the following we briefly
summarize miscellaneous considerations.

%-------------------------------%
\paradot{Comparison to Rademacher complexities}
%-------------------------------%
For a (binary) classification problem,
the rank \eqref{eqRank} of classifier $r$ can be re-formulated
as the probability that a randomly relabeled sample $\v y'$ behaves better than the actual $\v y$.
The more flexible $r$ is, the larger its rank is.
The Rademacher complexity \citep{Kol:01,BBL:02} of $r$
is the expectation of the difference between the misclassifying loss under
the actual $\v y$ and the misclassifying loss under a randomly relabeled sample $\v y'$.
The more flexible $r$ is, the larger its Rademacher complexity is.
Therefore, there is a close connection between LoRP and Rademacher complexities.
Model selection based on Rademacher complexities has
a number of attractive properties
and has been attracting many researchers,
thus it's worth discovering this connection.
Some results have been recently already obtained,
however, to keep the present paper not so long,
we decide to present the results in another paper.

%-------------------------------%
\paradot{Monte Carlo estimates for non-linear LoRP}
%-------------------------------%
For non-linear regression we did not present an efficient algorithm
for the loss rank/volume $\LR_r(\v y|\v x)$. The high-dimensional
volume $|V_r(L)|$ \eqref{defV} may be computed by Monte Carlo
algorithms. Normally $V_r(L)$ constitutes a small part of $\Y^n$, and
uniform sampling over $\Y^n$ is not feasible. Instead one should
consider two competing regressors $r$ and $r'$ and compute $|V\cap
V'|/|V|$ and $|V\cap V'|/|V'|$ by uniformly sampling from $V$ and
$V'$ respectively e.g.,\ with a Metropolis-type algorithm. Taking the
ratio we get $|V'|/|V|$ and hence the loss rank difference
$\LR_r-\LR_{r'}$, which is sufficient for LoRP. The usual tricks and
problems with sampling apply here too.

%-------------------------------%
\paradot{LoRP for hybrid model classes}
%-------------------------------%
LoRP is not restricted to model classes indexed by a
single integral ``complexity'' parameter, but may be applied more
generally to selecting among some (typically discrete) class of
models/regressors. For instance, the class could contain kNN {\em
and} polynomial regressors, and LoRP selects the complexity {\em
and} type of regressor (non-parametric kNN versus parametric
polynomials).

%-------------------------------%
\paradot{Generative versus discriminative LoRP}
%-------------------------------%
We have concentrated on counting $y$'s given fixed $x$, which
corresponds to discriminative learning. LoRP might equally well be
used for counting $(x,y)$, as alluded to in the introduction. This
would correspond to generative learning. Both regimes are used in
practice. See \cite{Liang:08} for some recent results on their
relative benefit, and further references.

%-------------------------------%
\paradot{Acknowledgement}
%-------------------------------%
We would like to thank two anonymous reviewers for their
detailed and helpful comments.
The second author would like to thank the SML$@$NICTA for supporting
a visit which led to the present paper.

%%%%%%%%%%%%%%%%%%%%%%%%%%%%%%%%%%%%%%%%%%%%%%%%%%%%%%%%%%%%%%%
\section*{Appendix: List of Abbreviations and Notations}\label{secApp}
%%%%%%%%%%%%%%%%%%%%%%%%%%%%%%%%%%%%%%%%%%%%%%%%%%%%%%%%%%%%%%%

AIC= Akaike Information Criterion.\\
BIC= Bayesian Information Criterion.\\
BMS= Bayesian Model Selection\\
kNN= k Nearest Neighbors.\\
LBFR= Linear Basis Function Regression.\\
LoRP= Loss Rank Principle.\\
LRC = Loss Rank Code.\\
MAP= Maximum a Posterior.\\
MDL= Minimum Description Length.\\
ML= Maximum Likelihood.\\
PML= Penalized Maximum Likelihood.\\
$D=\{(x_1,y_1),...,(x_n,y_n)\}$= observed data.\\
$\D=\{D\}$= set of all possible data $D$.\\
$\X\times\Y$=observation space.\\
$\v x=(x_1,...,x_n)$= vector of $x$-observations, similarly $\v y$.\\
$f:\X\to\Y$= functional dependence between $x$ and $y$.\\
$\F$= (``small'') class of functions $f$.\\
$\cal H$= class of stochastic hypotheses/models.\\
$r:\D\to\F$= regressor/model.\\
$\h y_i=r(x_i|D)$= $r$-estimate of $y_i$.\\
$\R$= (``small'') class of regressors/models.\\
$\v w\in\SetR^d$= parametrization of $\F_d$.\\
$\N_k(x)$= set of indices of the $k$ nearest neighbors of $x$ in $D$.\\
$L=\L_r(D)=\L(\v y,\v{\h y})$= empirical loss of $r$ on $D$.\\
$\Rank_r(L)=\#\{\v y'\in\Y^n : \L_r(\v y'|\v x)\leq L\}$= loss rank of $r$.\\
$V(L)$= volume of $D$ under $r$.\\
$\LR_r(\v y|\v x)$= log rank/volume of $D$.\\
$\LR_r^\a$= regularized $\LR_r$.\\
$d_{e\!f\!f}$= effective dimension.\\
$m_j(x,\v x)$= coefficients of linear regressor.\\
$M(\v x)$= linear regression matrix or ``hat" matrix.\\
$\log$= natural logarithm.\\
$a\leadsto b$: $a$ is replaced by $b$.

%%%%%%%%%%%%%%%%%%%%%%%%%%%%%%%%%%%%%%%%%%%%%%%%%%%%%%%%%%%%%%%
%         Bibliography        %
%%%%%%%%%%%%%%%%%%%%%%%%%%%%%%%%%%%%%%%%%%%%%%%%%%%%%%%%%%%%%%%
\addcontentsline{toc}{section}{\refname}
\bibliographystyle{alpha}
\begin{small}

\end{small}
\end{document}